\documentclass{article} 
\usepackage{iclr2024_conference,times}

\usepackage{amsmath,amsfonts,bm}

\def\eqref#1{equation~\ref{#1}}

\def\1{\bm{1}}

\DeclareMathAlphabet{\mathsfit}{\encodingdefault}{\sfdefault}{m}{sl}
\SetMathAlphabet{\mathsfit}{bold}{\encodingdefault}{\sfdefault}{bx}{n}

\usepackage[colorlinks=true,citecolor=.]{hyperref}
\usepackage{graphicx}
\usepackage{url}
\usepackage{wrapfig}
\usepackage{graphics}
\usepackage{booktabs}
\usepackage{subcaption}
\usepackage{multirow}
\usepackage{bbm}
\usepackage{amsmath,amsfonts,bm}
\usepackage{makecell}
\usepackage{array}
\usepackage[para]{threeparttable}
\usepackage{booktabs}
\usepackage{tabularx}
\usepackage{xspace}
\usepackage{hyperref}
\usepackage{tablefootnote}
\usepackage{adjustbox}
\usepackage[most]{tcolorbox}
\usepackage{colortbl}

\newtcolorbox{mybox}[2][]{
    colback=white,
    colframe=green!45,
    fonttitle=\bfseries,
    coltitle=black,
    sharp corners,
    title=#2,
    #1
}

\title{\raisebox{-0.1cm}{\includegraphics[width=0.500cm, height=0.637cm]{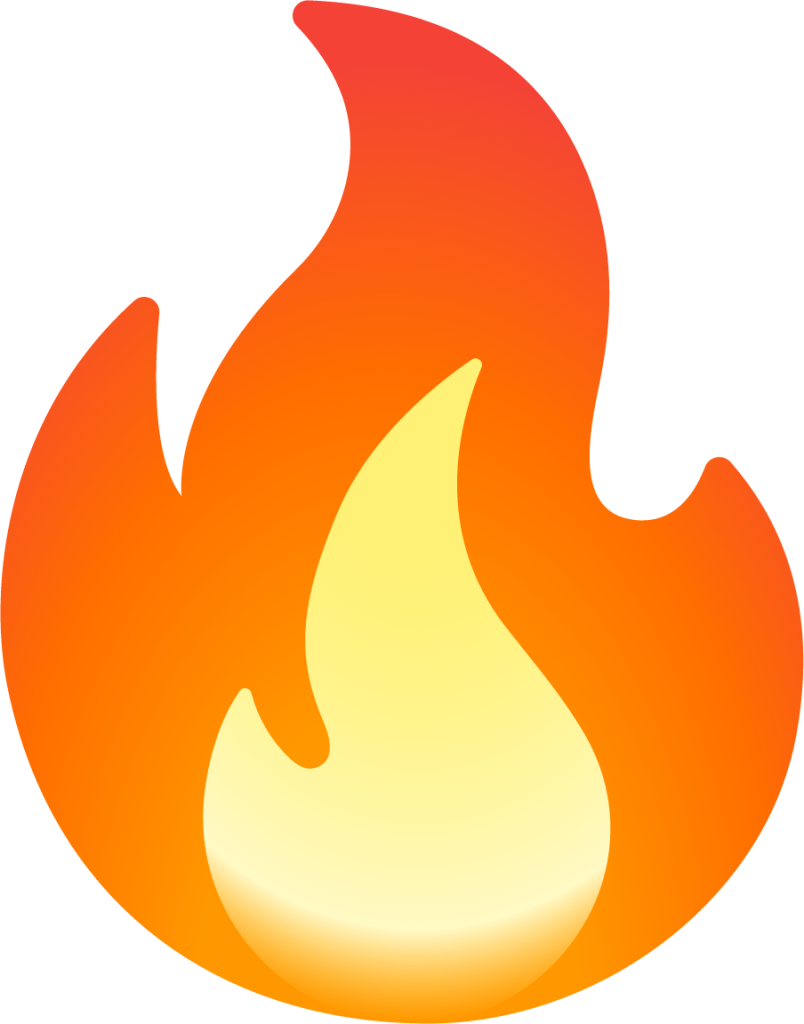}}~Prometheus: Inducing Fine-grained\\ Evaluation Capability in Language Models}

\author{Seungone Kim$^{1,2}$\thanks{denotes equal contribution. Work was done while Seungone was interning at NAVER AI Lab.}~~\thanks{Corresponding authors} \quad Jamin Shin$^{2,3*}$$\textbf{}^{\dagger}$ \quad Yejin Cho$^{1*}$$\textbf{}^{\dagger}$ \quad Joel Jang$^{4}$ \quad Shayne Longpre$^{5}$ \\ \textbf{Hwaran Lee}$^{2,3}$ \quad \textbf{Sangdoo Yun}$^{2,3}$ \quad \textbf{Seongjin Shin}$^{3}$ \quad \textbf{Sungdong Kim}$^{1,2,3}$ \\ \textbf{James Thorne}$^{1}$ \quad \textbf{Minjoon Seo}$^{1}$$\textbf{}^{\dagger}$\\
\\$^{1}$KAIST AI \quad $^{2}$NAVER AI Lab \quad $^{3}$NAVER Cloud \quad $^{4}$University of Washington \quad $^{5}$MIT\\ \\
\texttt{\{seungone, yejin\_cho, minjoon\}@kaist.ac.kr} \quad \texttt{jamin.shin@outlook.com}\\
}

\iclrfinalcopy 
\begin{document}

\maketitle

\begin{abstract}
Recently, using a powerful proprietary Large Language Model (LLM) (e.g., GPT-4) as an evaluator for long-form responses has become the de facto standard. 
However, for practitioners with large-scale evaluation tasks and custom criteria in consideration (e.g., child-readability), using proprietary LLMs as an evaluator is unreliable due to the closed-source nature, uncontrolled versioning, and prohibitive costs.
In this work, we propose \textsc{Prometheus}, a fully open-source LLM that is on par with GPT-4's evaluation capabilities when the appropriate reference materials (reference answer, score rubric) are accompanied.
We first construct the \textsc{Feedback Collection}, a new dataset that consists of 1K fine-grained score rubrics, 20K instructions, and 100K responses and language feedback generated by GPT-4. 
Using the \textsc{Feedback Collection}, we train \textsc{Prometheus}, a 13B evaluator LLM that can assess any given long-form text based on \textit{customized} score rubric provided by the user. 
Experimental results show that \textsc{Prometheus} scores a Pearson correlation of 0.897 with human evaluators when evaluating with 45 customized score rubrics, which is on par with GPT-4 (0.882), and greatly outperforms ChatGPT (0.392). Furthermore, measuring correlation with GPT-4 with 1222 customized score rubrics across four benchmarks (MT Bench, Vicuna Bench, Feedback Bench, Flask Eval) shows similar trends, bolstering \textsc{Prometheus}'s capability as an evaluator LLM. Lastly, \textsc{Prometheus} achieves the highest accuracy on two human preference benchmarks (HHH Alignment \& MT Bench Human Judgment) compared to open-sourced reward models explicitly trained on human preference datasets, highlighting its potential as an universal reward model. We open-source our code, dataset, and model \footnote{\url{https://kaistai.github.io/prometheus/}}.

\end{abstract}

\section{Introduction}



Evaluating the quality of machine-generated text has been a long-standing challenge in Natural Language Processing (NLP) and remains especially essential in the era of Large Language Models (LLMs) to understand their properties and behaviors~\citep{liang2022holistic,chang2023survey,zhong2023agieval,chia2023instructeval,holtzman2023generative}. 
Human evaluation has consistently been the predominant method, for its inherent reliability and capacity to assess nuanced and subjective dimensions in texts. 
In many situations, humans can naturally discern the most important factors of assessment, such as brevity, creativity, tone, and cultural sensitivities. 
On the other hand, conventional automated evaluation metrics (e.g., BLEU, ROUGE) cannot capture the depth and granularity of human evaluation~\citep{papineni2002bleu, lin2004rouge, zhang2019bertscore, krishna2021hurdles}.




Applying LLMs (e.g. GPT-4) as an evaluator has received substantial attention due to its potential parity with human evaluation~\citep{chiang2023can,dubois2023alpacafarm,alpaca_eval,liu2023gpteval,peng2023instruction,zheng2023judging,ye2023flask,min2023factscore}.
Initial investigations and observations indicate that, when aptly prompted, LLMs can emulate the fineness of human evaluations. However, while the merits of using proprietary LLMs as an evaluation tool are evident, there exist some critical disadvantages:
\begin{enumerate}
    \item \textbf{Closed-source Nature}: The proprietary nature of LLMs brings transparency concerns as internal workings are not disclosed to the broader academic community. Such a lack of transparency hinders collective academic efforts to refine or enhance its evaluation capabilities. Furthermore, this places fair evaluation, a core tenet in academia, under control of for-profit entity and raises concerns about neutrality and autonomy.
    \item \textbf{Uncontrolled Versioning}: Proprietary models undergo version updates that are often beyond the users' purview or control ~\citep{pozzobon2023challenges}. This introduces a reproducibility challenge. As reproducibility is a cornerstone of scientific inquiry, any inconsistency stemming from version changes can undermine the robustness of research findings that depend on specific versions of the model, especially in the context of evaluation.
    \item \textbf{Prohibitive Costs}: Financial constraints associated with LLM APIs are not trivial.
    For example, evaluating four LLMs variants across four sizes (ranging from 7B to 65B) using GPT-4 on 1000 evaluation instances can cost over \$2000.
    Such scaling costs can be prohibitive, especially for academic institutions or researchers operating on limited budgets.
\end{enumerate}

\begin{figure*}[t!]
\centering
    \includegraphics[width=1.0\linewidth]{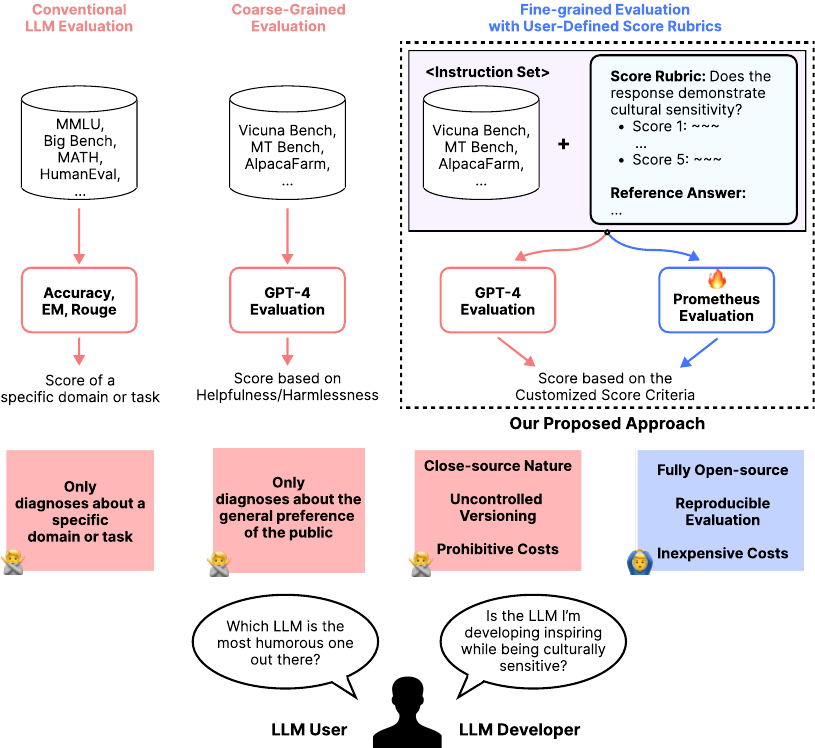}
    \caption{Compared to conventional, coarse-grained LLM evaluation, we propose a fine-grained approach that takes user-defined score rubrics as input.}
    \label{figure:main_figure}
\end{figure*}

Despite these limitations, proprietary LLMs such as GPT-4 are able to evaluate scores based on \textit{customized} score rubrics. Specifically, current resources are confined to \emph{generic}, single-dimensional evaluation metrics that are either too domain/task-specific (e.g. EM, Rouge) or coarse-grained (e.g. helpfulness/harmlessness~\citep{dubois2023alpacafarm,vicuna2023,liu2023gpteval} as shown in left-side of Figure~\ref{figure:main_figure}.
For instance, AlpacaFarm's~\citep{dubois2023alpacafarm} prompt gives a single definition of preference, asking the model to choose the model response that is \textit{generally} preferred. However, response preferences are subject to variation based on specific applications and values. 
In real-world scenarios, users may be interested in \textit{customized} rubric such as ``Which LLM generates responses that are playful and humorous'' or ``Which LLM answers with particularly care for cultural sensitivities?'' 
Yet, in our initial experiments, we observe that even the largest open-source LLM (70B) is insufficient to evaluate based on a customized score rubric compared to proprietary LLMs.

To this end, we propose \textsc{Prometheus}, a 13B LM that aims to induce fine-grained evaluation capability of GPT-4, while being open-source, reproducible, and inexpensive. We first create the \textsc{Feedback Collection}, a new dataset that is crafted to encapsulate diverse and fine-grained user assessment score rubric that represent realistic user demands (example shown in Figure~\ref{figure:feedback_collection}). 
We design the \textsc{Feedback Collection} with the aforementioned consideration in mind, encompassing thousands of unique preference criteria encoded by a user-injected score rubric.
Unlike prior feedback datasets~\citep{selfee2023,wang2023shepherd}, it uses \emph{custom}, not \emph{generic} preference score rubric, to train models to flexibly generalize to practical and diverse evaluation preferences. Also, to best of our knowledge, we are first to explore the importance of including various reference materials -- particularly the `Reference Answers' -- to effectively induce fine-grained evaluation capability. 

We use the \textsc{Feedback Collection} to fine-tune Llama-2-Chat-13B in creating \textsc{Prometheus}. On 45 customized score rubrics sampled across three test sets (MT Bench, Vicuna Bench, Feedback Bench), \textsc{Prometheus} obtains a Pearson correlation of 0.897 with human evaluators, which is similar with GPT-4 (0.882), and has a significant gap with GPT-3.5-Turbo (0.392). Unexpectely, when asking human evaluators to choose a feedback with better quality in a pairwise setting, \textsc{Prometheus} was preferred over GPT-4 in 58.67\% of the time, while greatly outperformed GPT-3.5-Turbo with a 79.57\% win rate. Also, when measuring the Pearson correlation with GPT-4 evaluation across 1222 customized score rubrics across 4 test sets (MT Bench, Vicuna Bench, Feedback Bench, Flask Eval), \textsc{Prometheus} showed higher correlation compared to GPT-3.5-Turbo and Llama-2-Chat 70B. Lastly, when testing on 2 unseen human preference datasets (MT Bench Human Judgments, HHH Alignment), \textsc{Prometheus} outperforms two state-of-the-art reward models and GPT-3.5-Turbo, highlighting its potential as an universal reward model.

Our contributions are summarized as follows:
\begin{itemize}
    \item We introduce the \textsc{Feedback Collection} dataset specifically designed to train an evaluator LM. Compared to previous feedback datasets, it includes customized scoring rubrics and reference answers in addition to the instructions, responses, and feedback.
   \item We train \textsc{Prometheus}, the first open-source LLM specialized for fine-grained evaluation that can generalize to diverse, real-world scoring rubrics beyond a single-dimensional preference such as helpfulness and harmlessness. 
   \item We conduct extensive experiments showing that by appending reference materials (reference answers, fine-grained score rubrics) and fine-tuning on feedback, we can induce evaluation capability into language models. \textsc{Prometheus} shows high correlation with human evaluation, GPT-4 evaluation in absolute scoring settings, and also shows high accuracy in ranking scoring settings.
\end{itemize}


\section{Related Work}

\paragraph{Reference-based text evaluation}
Previously, model-free scores that evaluate machine-generated text based on a golden candidate reference such as BLEU~\citep{papineni2002bleu} and ROUGE~\citep{lin-2004-rouge} scores were used as the dominant approach. However, \citet{krishna2021hurdles} reported limitations in reference-based metrics, such as ROUGE, observing that they are not reliable for evaluation. In recent years, model-based approaches have been widely adopted such as BERTScore~\citep{zhang2019bertscore}, BLEURT~\citep{Sellam2020BLEURTLR}, and BARTScore~\citep{yuan2021bartscore} which are able to capture \textit{semantic} information rather than only evaluating on \textit{lexical} components.

\paragraph{LLM-based text evaluation}

Recent work has used GPT-4 or a fine-tuned critique LLM as an evaluator along a single dimension of ``preference''~\citep{chiang2023can,alpaca_eval,dubois2023alpacafarm,zheng2023judging,liu2023gpteval}. 
For instance, AlpacaFarm~\citep{dubois2023alpacafarm} asks the model to select ``which response is better based on your judgment and based on your own preference'' Another example is recent work that showed ChatGPT can outperform crowd-workers for text-annotation tasks~\citep{gilardi2023chatgpt, chiang2023can}.
\citet{wang2023pandalm} introduced PandaLM, a fine-tuned LLM to evaluate the generated text and explain its reliability on various preference datasets. 
Similarly, \citet{selfee2023} and \citet{wang2023shepherd} create critique LLMs.
However, the correct preference is often subjective and depends on applications, cultures, and objectives, where degrees of brevity, formality, honesty, creativity, and political tone, among many other potentially desirable traits that may vary~\citep{chiang2023can}. 
While GPT-4 is unreliable due to its close-source nature, uncontrolled versioning, and prohibitive costs, it was the \textit{only} option explored for fine-grained and customized evaluation based on the score rubric~\citep{ye2023flask}.
On the contrary, we train, to best of our knowledge, the first evaluator sensitive to thousands of unique preference criteria, and show it significantly outperforms uni-dimensional preference evaluators in a number of realistic settings. Most importantly, compared to previous work, we strongly argue the importance of appending reference materials (score rubric and reference answer) in addition to fine-tuning on the feedback in order to effectively induce fine-grained evaluation capability.


\section{The Feedback Collection Dataset}\label{section:3}

While previous work has demonstrated that fine-tuning on feedback is effective for improving LMs to function as a critique~\citep{selfee2023,wang2023shepherd}, the datasets used in previous work are not directly applicable for improving LMs to function as a fine-grained \textit{evaluator}. We thus introduce the \textsc{Feedback Collection}, a new dataset for the sole purpose of fine-tuning an open-sourced evaluator LLM. Our 4 main considerations during dataset construction are: (1) including as many reference materials (i.e. reference answer, and scoring rubric) as possible, (2) maintaining a uniform length among the reference answers for each score (1 to 5) to prevent undesired length bias, (3) maintaining a uniform score distribution to prevent undesired decision bias, and (4) limiting the scope of our instructions and responses to realistic situations where a user is interacting with a LLM. 

\begin{figure*}[t!]
\centering
    \includegraphics[width=0.99\linewidth]{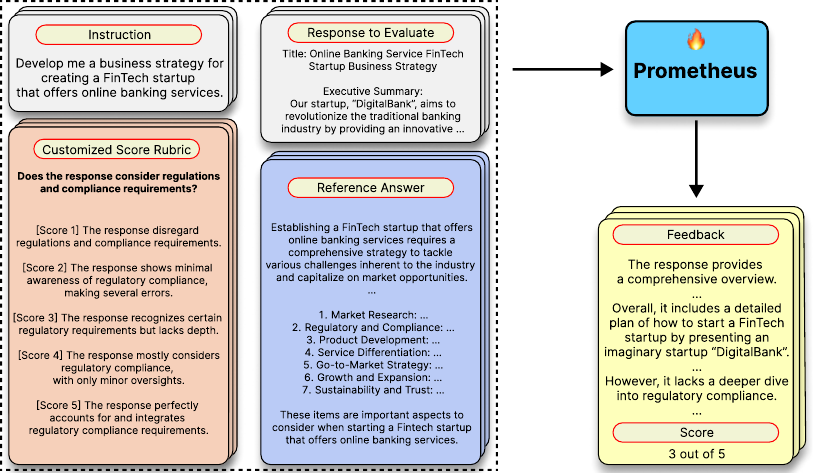}
    \caption{The individual components of the \textsc{Feedback Collection}. By adding the appropriate reference materials (Score Rubric and Reference Answer) and training on GPT-4's feedback, we show that we could obtain a strong open-source evaluator LM.}
    \label{figure:feedback_collection}
\end{figure*}


Taking these into consideration, we construct each instance within the \textsc{Feedback Collection} to encompass four components for the \textit{input} (instruction, response to evaluate, customized score rubric, reference answer) and two components in the \textit{output} (feedback, score). An example of an instance is shown in Figure \ref{figure:feedback_collection} and the number of each component is shown in Table~\ref{tab:train-sets}.

The four components for the input are as follows:
\begin{enumerate}
    \item \textbf{Instruction}: An instruction that a user would prompt to an arbitrary LLM.
    \item \textbf{Response to Evaluate}: A response to the instruction that the evaluator LM has to evaluate.
    \item \textbf{Customized Score Rubric}: A specification of novel criteria decided by the user. The evaluator should focus on this aspect during evaluation. The rubric consists of (1) a description of the criteria and (2) a description of each scoring decision (1 to 5).
    \item \textbf{Reference Answer}: A reference answer that would receive a score of 5. Instead of requiring the evaluator LM to solve the instruction, it enables the evaluator to use the mutual information between the reference answer and the response to make a scoring decision.
\end{enumerate}

The two components for the output are as follows: 
\begin{enumerate}
    \item \textbf{Feedback}: A rationale of why the provided response would receive a particular score. This is analogous to Chain-of-Thoughts (CoT), making the evaluation process interpretable.
    \item \textbf{Score}: An integer score for the provided response that ranges from 1 to 5.
\end{enumerate}

\begin{table*}[t]
\centering
\caption{Information about our training dataset \textsc{Feedback Collection}. Note that there are 20 instructions accompanied for each score rubric, leading to a total number of 20K. Also, there is a score 1-5 response and feedback for each instruction, leading to a total number of 100K.}
\resizebox{\textwidth}{!}{%
\begin{tabular}{@{}ccccc@{}}
\toprule
\textbf{Evaluation Mode} &
  \textbf{Data} &
  \textbf{\# Score Rubrics} &
  \textbf{\# Instructions \&  Reference Answer} &
  \textbf{\# Responses \&  Feedback} \\ \midrule
\multirow{2}{*}{\textbf{Absolute Evaluation}} &
  \multirow{2}{*}{\textsc{Feedback Collection}} &
  \multirow{2}{*}{\begin{tabular}[c]{@{}c@{}}1K (Fine-grained \&\\ Customized) \end{tabular}} &
  \multirow{2}{*}{\begin{tabular}[c]{@{}c@{}}Total 20K\\ (20 for each score rubric) \end{tabular}} &
  \multirow{2}{*}{\begin{tabular}[c]{@{}c@{}}Total 100K(5 for each instruction;\\20K for each score within 1-5) \end{tabular}} \\
   \\ \bottomrule
\end{tabular}%
}
\label{tab:train-sets}
\end{table*}

\begin{figure*}[t!]
\centering
    \includegraphics[width=0.9\linewidth]{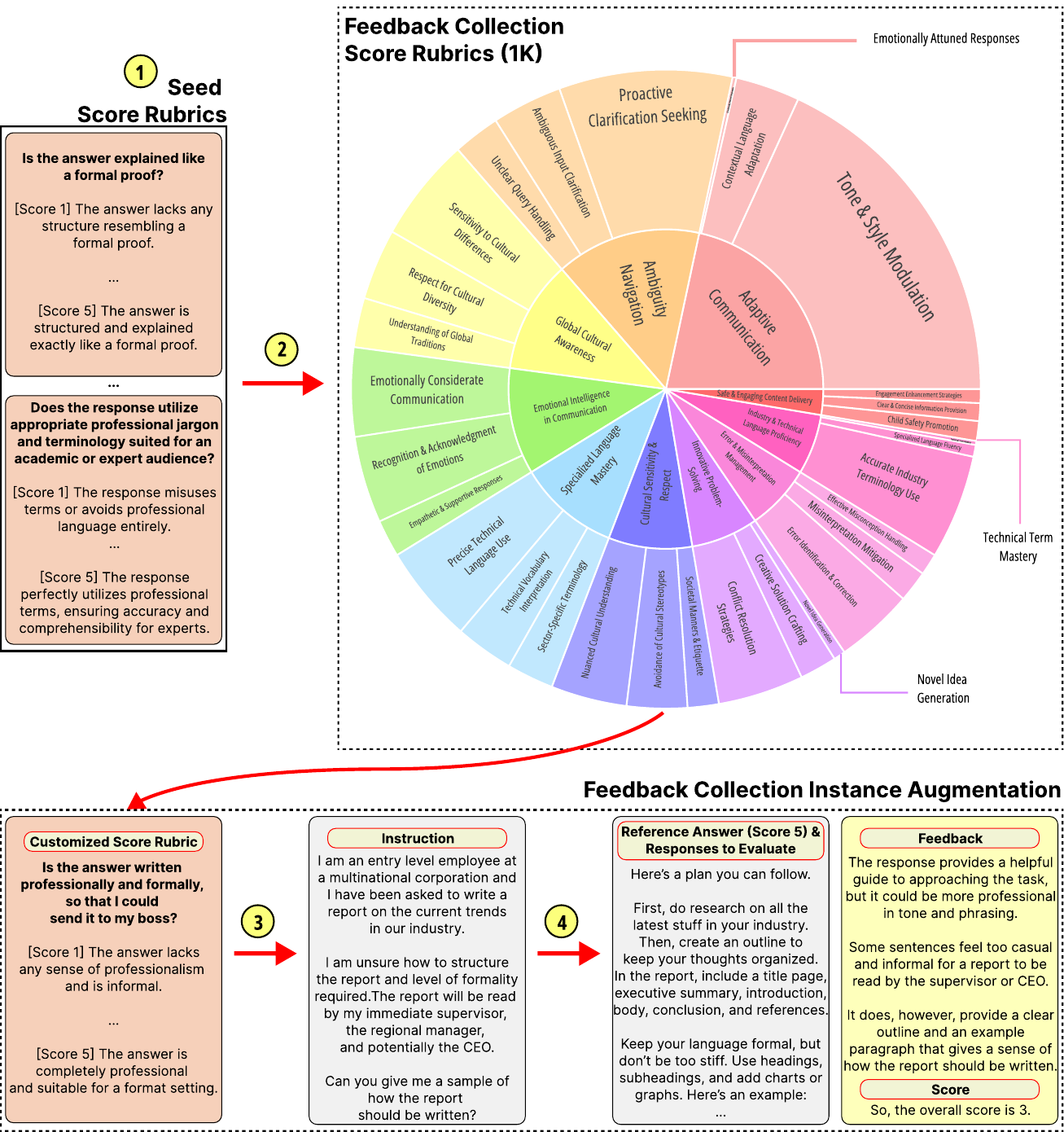}
    \caption{Overview of the augmentation process of the \textsc{Feedback Collection}. The keywords included within the score rubrics of the \textsc{Feedback Collection} is also displayed.}
    \label{figure:fc-augmentation}
\end{figure*}

Each instance has an accompanying scoring rubric and reference answer upon the instruction in order to include as much reference material as possible. Also, we include an equal number of 20K instances for each score in the range of 1 to 5, preventing undesired decision bias while training the evaluator LLM. A detailed analysis of the \textsc{Feedback Collection} dataset is in Appendix~\ref{appendix:feedback_collection}.

\subsection{Dataset Construction Process}
\label{subsec:data_collection}

We construct a large-scale \textsc{Feedback Collection} dataset by prompting GPT-4. Specifically, the collection process consists of (1) the curation of 50 initial seed rubrics, (2) the expansion of 1K new score rubrics through GPT-4, (3) the augmentation of realistic instructions, and (4) the augmentation of the remaining components in the training instances (i.e. responses including the reference answers, feedback, and scores). Figure~\ref{figure:fc-augmentation} shows the overall augmentation process. 

\paragraph{Step 1: Creation of the Seed Rubrics}
We begin with the creation of a foundational seed dataset of scoring rubrics. Each author curates a detailed and fine-grained scoring rubric that each personnel considers pivotal in evaluating outputs from LLMs. This results in an initial batch of 50 seed rubrics.  

\paragraph{Step 2: Augmenting the Seed Rubrics with GPT-4}
Using GPT-4 and our initial seed rubrics, we expand the score rubrics from the initial 50 to a more robust and diverse set of 1000 score rubrics. Specifically, by sampling 4 random score rubrics from the initial seed, we use them as demonstrations for in-context learning (ICL), and prompt GPT-4 to brainstorm a new novel score rubric. Also, we prompt GPT-4 to paraphrase the newly generated rubrics in order to ensure \textsc{Prometheus} could generalize to the similar score rubric that uses different words. We iterate the brainstorming $\rightarrow$ paraphrasing process for 10 rounds. The detailed prompt used for this procedure is in Appendix~\ref{appendix:prompt_augmentation}.

\paragraph{Step 3: Crafting Novel Instructions related to the Score Rubrics}
With a comprehensive dataset of 1000 rubrics at our disposal, the subsequent challenge was to craft pertinent training instances. For example, a score rubric asking ``Is it formal enough to send to my boss'' is not related to a math problem. Considering the need for a set of instructions closely related to the score rubrics, we prompt GPT-4 to generate 20K unique instructions that are highly relevant to the given score rubric.

\paragraph{Step 4: Crafting Training Instances}
Lastly, we sequentially generate a response to evaluate and corresponding feedback by prompting GPT-4 to generate each component that will get a score of $i$ (1 $\leq i \leq$ 5). This leads to 20 instructions for each score rubric, and 5 responses \& feedback for each instruction. To eliminate the effect of decision bias when fine-tuning our evaluator LM, we generate an equal number of 20K responses for each score. Note that for the response with a score of 5, we generate two distinctive responses so we could use one of them as an input (reference answer).

\subsection{Fine-tuning an Evaluator LM}
Using the \textsc{Feedback Collection} dataset, we fine-tune Llama-2-Chat (7B \& 13B) and obtain \textsc{Prometheus} to induce fine-grained evaluation capability. Similar to Chain-of-Thought Fine-tuning~\citep{ho2022large,kim2023cot}, we fine-tune to sequentially generate the feedback and then the score. We highlight that it is important to include a phrase such as `\texttt{[RESULT]}' in between the feedback and the score to prevent degeneration during inference. We include the details of fine-tuning, inference, and ablation experiments (reference materials, base model) in Appendix~\ref{appendix:hyperparameter_details}.

\section{Experimental Setting: Evaluating an Evaluator LM}

In this section, we explain our experiment setting, including the list of experiments, metrics, and baselines that we use to evaluate \textit{fine-grained} evaluation capabilities of an evaluator LLM. Compared to measuring the instruction-following capability of a LLM, it is not straightforward to directly measure the capability to evaluate. Therefore, we use human evaluation and GPT-4 evaluation as a standard and measure how similarly our evaluator model and baselines could closely simulate them. We mainly employ two types of evaluation methods: \textit{Absolute Grading} and \textit{Ranking Grading}. Detailed information on the datasets used for the experiment is included in Table~\ref{tab:eval-sets}.

\subsection{List of Experiments and Metrics}

\paragraph{Absolute Grading}
We first test in an Absolute Grading setting, where the evaluator LM should generate a feedback and score within the range of 1 to 5 given an instruction, a response to evaluate, and reference materials (as shown in Figure~\ref{figure:feedback_collection}). Absolute Grading is challenging compared to Ranking Grading since the evaluator LM does not have access to an opponent to compare with and it is required to provide a score \textit{solely} based on its internal decision. Yet, it is more practical for users since it relieves the need to prepare an opponent to compare with during evaluation.

We mainly conduct three experiments in this setting: (1) measuring the correlation with human evaluators (Section~\ref{subsec:5.1}), (2) comparing the quality of the feedback using human evaluation (Section~\ref{subsec:5.1}), and (3) measuring the correlation with GPT-4 evaluation (Section~\ref{subsec:5.2}). For the experiments that measure the correlation, we use 3 different correlation metrics: \textbf{Pearson}, \textbf{Kdendall-Tau}, and \textbf{Spearman}. For measuring the quality of the generated feedback, we conduct a \textbf{pairwise comparison} between the feedback generated by \textsc{Prometheus}, GPT-3.5-Turbo, and GPT-4, asking human evaluators to choose \textit{which} has better quality and \textit{why} they thought so. Specifically, we recruited 9 crowdsource workers and split them into three groups: \textsc{Prometheus} vs GPT-4, \textsc{Prometheus} vs ChatGPT, and GPT-4 vs ChatGPT. The annotators are asked to answer the following three questions:
\begin{enumerate}
    \item What score would you give to the response based on the given score rubric?
    \item Among the two Feedback, which is better for critiquing the given response?
    \item Why did you reject that particular feedback?
\end{enumerate}

\begin{table*}[t]
\centering
\caption{Information about the datasets we use to test evaulator LMs. Note that \textsc{Feedback Bench} is a dataset that is crafted with the exact same procedure as the \textsc{Feedback Collection} as explained in Section~\ref{subsec:data_collection}. We include additional analysis of \textsc{Feedback Bench} in Appendix~\ref{appendix:feedback_bench}. Simulated GPT-4 $\dagger$ denotes GPT-4 prompted to write a score of $i$ $(1 \leq i \leq 5$) during augmentation.}
\resizebox{\textwidth}{!}{%
\begin{tabular}{@{}cccc@{}}
\toprule
\textbf{Evaluation Mode} &
  \textbf{Evaluation Data} &
  \textbf{Source / Types of Score Rubric} &
  \textbf{Response LMs} \\ \midrule
\multirow{10}{*}{\textbf{Absolute Evaluation}} &
  \multirow{2}{*}{\textsc{Feedback Bench} (Seen Rubric)} &
  \multirow{2}{*}{1K Machine Generated} &
  \multirow{4}{*}{Simluated GPT-4 $\dagger$} \\
 &
   &
   &
   \\
 &
  \multirow{2}{*}{\textsc{Feedback Bench} (Unseen Rubric)} &
  \multirow{2}{*}{50 Hand Crafted} &
   \\
 &
   &
   &
   \\ \cmidrule(l){2-4} 
 &
  \multirow{2}{*}{Vicuna Bench} &
  \multirow{4}{*}{80 Hand Crafted} &
  \multirow{4}{*}{\begin{tabular}[c]{@{}c@{}}WizardLM, Vicuna,\\ Llama2-Chat, ChatGPT\end{tabular}} \\
 &
   &
   &
   \\
 &
  \multirow{2}{*}{MT Bench} &
   &
   \\
 &
   &
   &
   \\ \cmidrule(l){2-4} 
 &
  \multirow{2}{*}{Flask Eval} &
  \multirow{2}{*}{\begin{tabular}[c]{@{}c@{}}Logical Thinking (3), Background Knowledge (2)\\ Problem Handling (4), User Alignment (3)\end{tabular}} &
  \multirow{2}{*}{Alpaca, Vicuna, Bard, ChatGPT} \\
 &
   &
   &
   \\ \midrule
\multirow{4}{*}{\textbf{Ranking Evaluation}} &
  \multirow{2}{*}{MT Bench Human Judgments} &
  \multirow{2}{*}{Human Preference (Helpfulness)} &
  \multirow{2}{*}{\begin{tabular}[c]{@{}c@{}}Alpaca, Llama, Vicuna, ChatGPT, \\ Claude-v1, GPT-4 \end{tabular}} \\
 &
   &
   &
   \\ \cmidrule(l){2-4} 
 &
  \multirow{2}{*}{HHH Alignment} &
  \multirow{2}{*}{\begin{tabular}[c]{@{}c@{}}Helpfulness, Harmlessness, \\ Honesty, Other\end{tabular}} &
  \multirow{2}{*}{Human Annotation} \\
 &
   &
   &
   \\ \bottomrule
\end{tabular}%
}
\label{tab:eval-sets}
\end{table*}

We use the following four benchmarks to measure the correlation with human evaluation and GPT-4 evaluation. Note that \textsc{Feedback Bench} is a dataset generated with the same procedure as the \textsc{Feedback Collection}, and is divided into two subsets (Seen Rubric and Unseen Rubric).
\begin{itemize}
    \item \textbf{\textsc{Feedback Bench}}: The \textbf{Seen Rubric} subset shares the same 1K score rubrics with the \textsc{Feedback Collection} across 1K instructions (1 per score rubric). The \textbf{Unseen Rubric} subset also consists of 1K new instructions but with 50 new score rubrics that are generated the same way as the training set. Details are included in Appendix~\ref{appendix:feedback_bench}.
    \item \textbf{Vicuna Bench}: We adapt the 80 test prompt set from Vicuna~\citep{vicuna2023} and hand-craft customized score rubrics for each test prompt. In order to obtain reference answers, we concatenate the hand-crafted score rubric and instruction to prompt GPT-4.
    \item \textbf{MT Bench}: We adapt the 80 test prompt set from MT Bench~\citep{zheng2023judging}, a multi-turn instruction dataset. We hand-craft customized score rubrics and generate a reference answer using GPT-4 for each test prompt as well. Note that we only use the last turn of this dataset for evaluation, providing the previous dialogue as input to the evaluator LM.
    \item \textbf{FLASK Eval}: We adapt the 200 test prompt set from FLASK~\citep{ye2023flask}, a fine-grained evaluation dataset that includes multiple conventional NLP datasets and instruction datasets. We use the 12 score rubrics (that are relatively coarse-grained compared to the 1K score rubrics used in the \textsc{Feedback Collection}) such as Logical Thinking, Background Knowledge, Problem Handling, and User Alignment.
\end{itemize}

\paragraph{Ranking Grading}
To test if an evaluator LM trained only on Absolute Grading could be utilized as a universal reward model based on \textit{any} criteria, we use existing human preference benchmarks and use \textbf{accuracy} as our metric (Section~\ref{subsec:5.3}). Specifically, we check whether the evaluator LM could give a higher score to the response that is preferred by human evaluators. The biggest challenge of employing an evaluator LM trained in an Absolute Grading setting and testing it on Ranking Grading was that it could give the same score for both candidates. Therefore, we use a temperature of 1.0 when evaluating each candidate independently and iterate until there is a winner. Hence, it's noteworthy that the settings are not exactly fair compared to other ranking models. This setting is NOT designed to claim SOTA position in these benchmarks, but is conducted only for the purpose of checking whether an evaluator LM trained in an Absolute Grading setting could also generalize in a Ranking Grading setting according to \textit{general} human preference. Also, in this setting, we do not provide a reference answer to check whether \textsc{Prometheus} could function as a reward model. We use the following two benchmarks to measure the accuracy with human preference datasets:
\begin{itemize}
    \item \textbf{MT Bench Human Judgement}: This data is another version of the aforementioned MT Bench~\citep{zheng2023judging}. Note that it includes a tie option as well and does not require iterative inference to obtain a clear winner. We use Human Preference as our criteria.
    \item \textbf{HHH Alignment}: Introduced by Anthropic~\citep{askell2021general}, this dataset (221 pairs) is one of the most widely chosen reward-model test-beds that measures preference accuracy in Helpfulness, Harmlessness, Honesty, and in General (Other) among two response choices.
\end{itemize}

\subsection{Baselines}
The following list shows the baselines we used for comparison in the experiments:
\begin{itemize}
    \item \textsc{Llama2-Chat-\{7,13,70\}B}~\citep{touvron2023llama2}: The base model of \textsc{Prometheus} when fine-tuning on the \textsc{Feedback Collection}. Also, it is considered as the best option among the open-source LLMs, which we use as an evaluator in this work.
    \item \textsc{LLama-2-Chat-13B + Coarse}: To analyze the effectiveness of training on thousands of fine-grained score rubrics, we train a comparing model only using 12 coarse-grained score rubrics from \citet{ye2023flask}. Detailed information on this model is in Appendix~\ref{appendix:coarse_grained}.
    \item \textsc{GPT-3.5-turbo-0613}: Proprietary LLM that offers a cheaper price when employed as an evaluator LLM. While it is relatively inexpensive compared to GPT-4, it still has the issue of uncontrolled versioning and close-source nature.
    \item \textsc{GPT-4-\{0314,0613, Recent\}}: One of the most powerful proprietary LLM that is considered the main option when using LLMs as evaluators. Despite its reliability as an evaluator LM due to its superior performance, it has several issues of prohibitive costs, uncontrolled versioning, and close-source nature.
    \item \textsc{StanfordNLP Reward Model}\footnote{\url{https://huggingface.co/stanfordnlp/SteamSHP-flan-t5-xl}}: One of the state-of-the-art reward model directly trained on multiple human preference datasets in a ranking grading setting.
    \item \textsc{ALMOST Reward Model}~\citep{kim2023aligning}: Another state-of-the-art reward model trained on synthetic preference datasets in a ranking grading setting.
\end{itemize}

\section{Experimental Results}
\label{sec:results}


\begin{figure*}[t!]
\centering
    \includegraphics[width=0.8\linewidth]{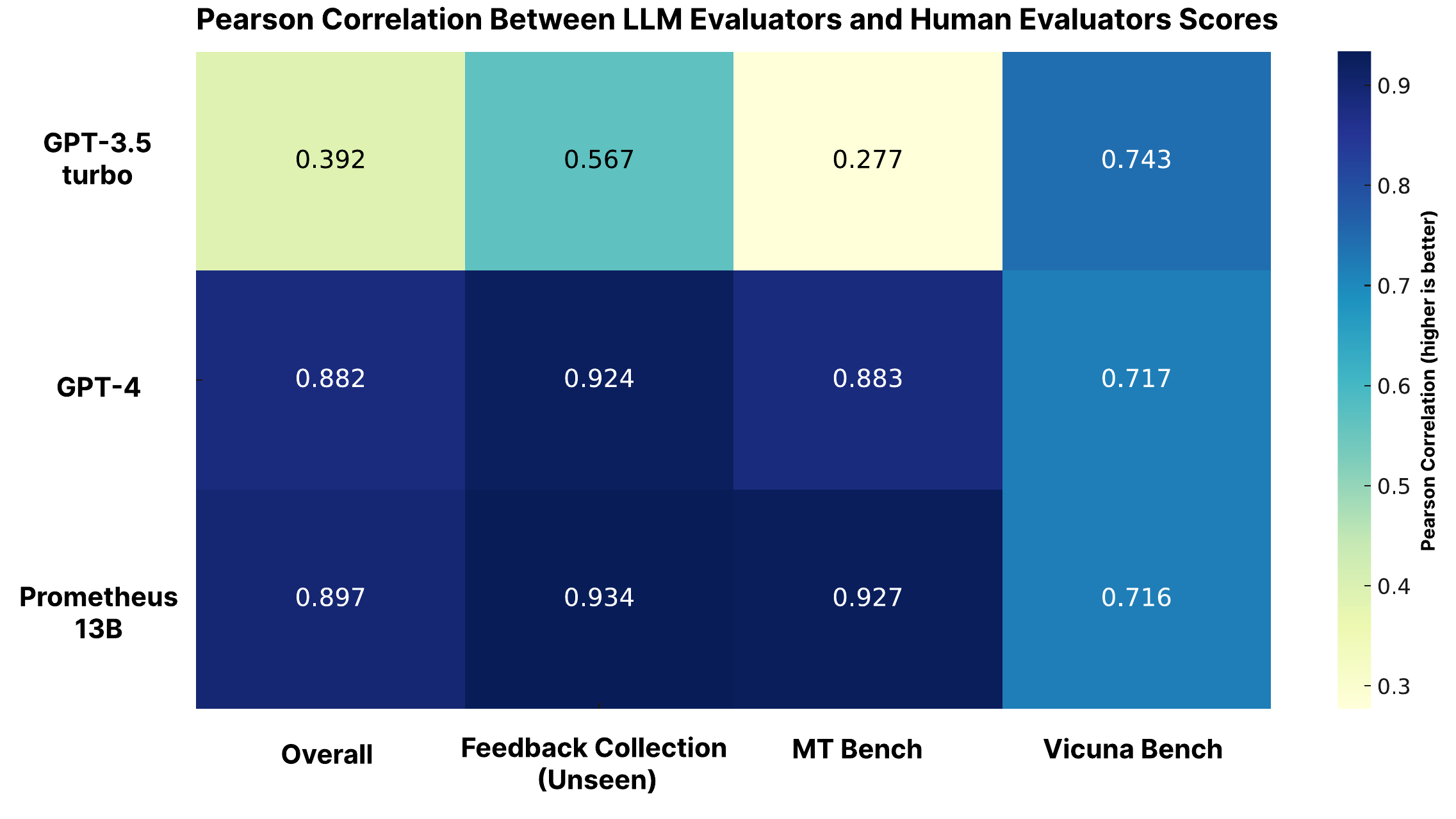}
    \caption{The Pearson correlation between scores from human annotators and the score from GPT-3.5-Turbo, \text{Prometheus}, and GPT-4 on 45 customized score rubrics from the Feedback Bench, Vicuna Bench, and MT Bench. \textsc{Prometheus} shows a high correlation with human evaluators.}
    \label{figure:human_eval_scores}
\end{figure*}

\begin{figure*}[t!]
\centering
    \includegraphics[width=0.8\linewidth]{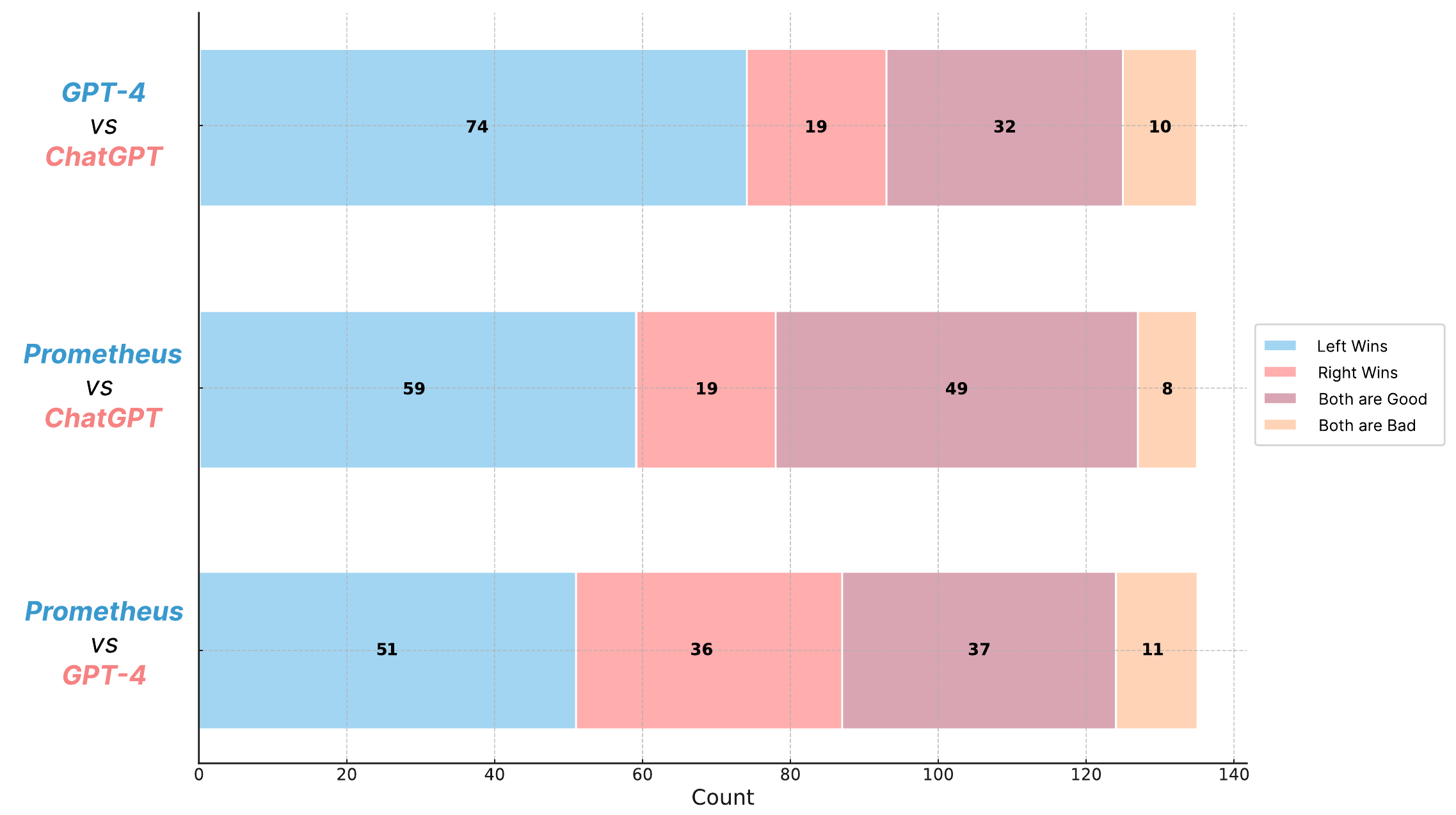}
    \caption{Pairwise comparison of the quality of the feedback generated by GPT-4, \textsc{Prometheus} and GPT-3.5-Turbo. Annotators are asked to choose which feedback is better at assessing the given response. \textsc{Prometheus} shows a win-rate of 58.62\% over GPT-4 and 79.57\% over GPT-3.5-Turbo.}
    \label{figure:human_eval_feedback}
\end{figure*}

\begin{figure*}[t!]
\centering
    \includegraphics[width=0.9\linewidth]{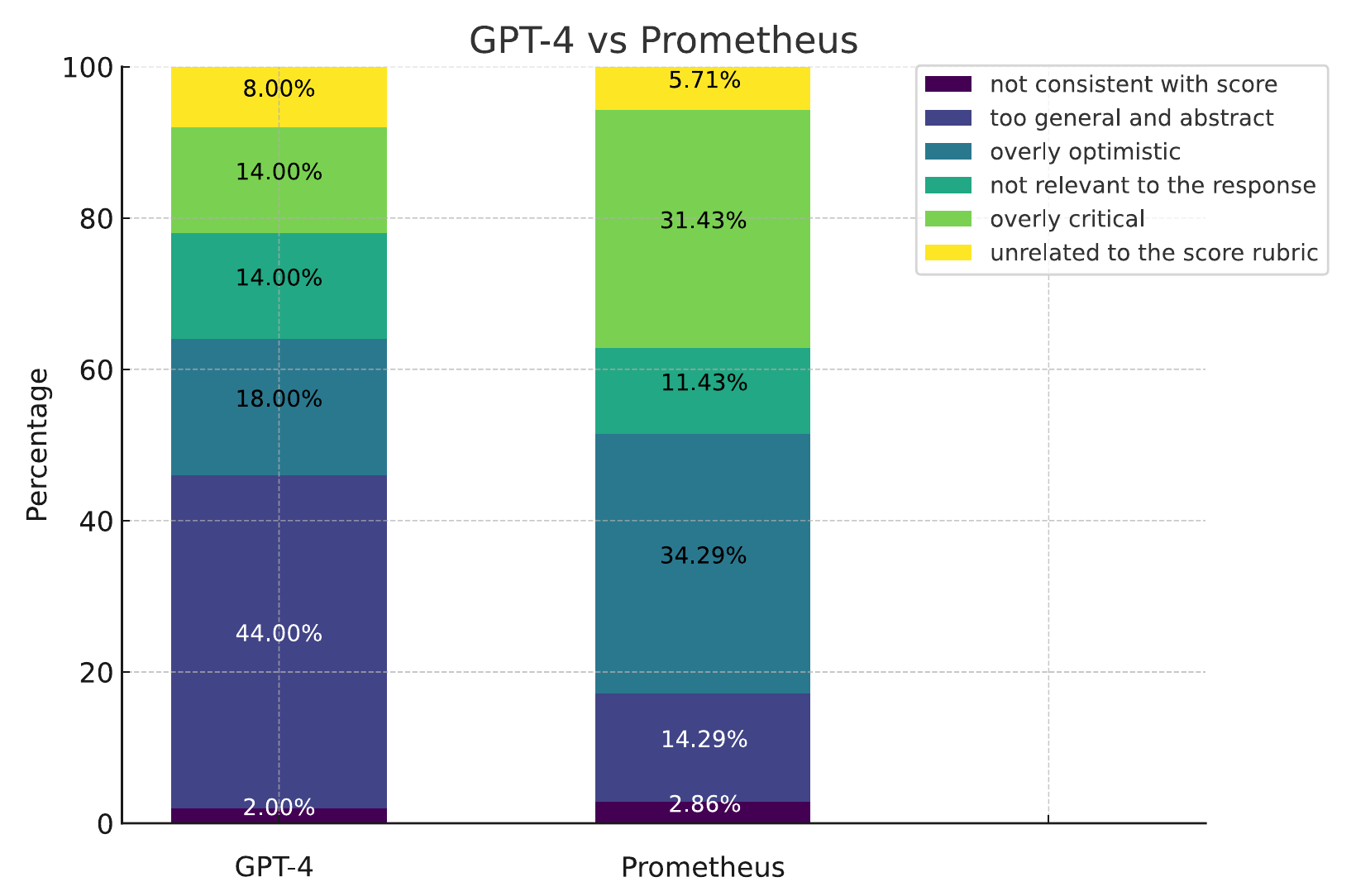}
    \caption{The reason why GPT-4's or Prometheus's feedback was not chosen over the other. \textsc{Prometheus} generates less abstract and general feedback, but tends to write overly critical ones.}
    \label{figure:human_eval_why_feedback_gpt4_prometheus}
\end{figure*}

\subsection{Can \textsc{Prometheus} Closely Simulate Human Evaluation?}\label{subsec:5.1}
\paragraph{Correlation with Human Scoring} We first compare the correlation between human annotators and our baselines using 45 instances each with an unique customized score rubric, namely the \textsc{Feedback Bench} (Unseen Score Rubric subset), MT Bench~\citep{zheng2023judging}, and Vicuna Bench~\citep{vicuna2023}. The results are shown in Figure~\ref{figure:human_eval_scores}, showing that \textsc{Prometheus} is on par with GPT-4 across all the three evaluation datasets, where \textsc{Prometheus} obtains a 0.897 Pearson correlation, GPT-4 obtains 0.882, and GPT-3.5-Turbo obtains 0.392. 

\paragraph{Pairwise Comparison of the Feedback with Human Evaluation} To validate the effect of whether \textsc{Prometheus} generates helpful/meaningful feedback in addition to its scoring decision, we ask human annotators to choose a better feedback. The results are shown in Figure~\ref{figure:human_eval_feedback}, showing that \textsc{Prometheus} is preferred over GPT-4 58.62\% of the times, and over GPT-3.5-Turbo 79.57\% of the times. This shows that \textsc{Prometheus}'s feedback is also meaningful and helpful.

\paragraph{Analysis of Why Prometheus's Feedback was Preferred} In addition to a pairwise comparison of the feedback quality, we also conduct an analysis asking human annotators to choose why they preferred one feedback over the other by choosing at least one of the comprehensive 6 options (``rejected feedback is not consistent with its score'' / ``too general and abstract'' / ``overly optimistic'' / ``not relevant to the response'' / ``overly critical'' / ``unrelated to the score rubric''). In Figure~\ref{figure:human_eval_why_feedback_gpt4_prometheus}, we show the percentage of why each evaluator LLM (GPT-4 and \textsc{Prometheus}) was rejected. It shows that while GPT-4 was mainly not chosen due to providing general or abstract feedback, \textsc{Prometheus} was mainly not chosen because it was too critical about the given response. Based on this result, we conclude that whereas GPT-4 tends to be more neutral and abstract, \textsc{Prometheus} shows a clear trend of expressing its opinion of whether the given response is good or not. We conjecture this is an effect of directly fine-tuning \textsc{Prometheus} to ONLY perform fine-grained evaluation, essentially converting it to an evaluator rather than a generator. We include (1) additional results of analyzing ``\textsc{Prometheus} \textit{vs} GPT-3.5-Turbo'' and ``GPT-4 \textit{vs} GPT-3.5-Turbo'' in Appendix~\ref{appendix:comparison_experiment} and (2) a detailed explanation of the experimental setting of human evaluation in Appendix~\ref{appendix:human_eval_details_last}.

\subsection{Can \textsc{Prometheus} Closely Simulate GPT-4 Evaluation?}\label{subsec:5.2}

\paragraph{Correlation with GPT-4 Scoring} We compare the correlation between GPT-4 evaluation and our baselines using 1222 score rubrics across 2360 instances from the \textsc{Feedback Bench} (Seen and Unseen Score Rubric Subset), Vicuna Bench~\citep{vicuna2023}, MT Bench~\citep{zheng2023judging}, and Flask Eval~\citep{ye2023flask} in an absolute grading scheme. Note that for the \textsc{Feedback Bench}, we measure the correlation with the scores \textbf{augmented} from GPT-4-0613, and for the other 3 datasets, we measure the correlation with the scores acquired by \textbf{inferencing} GPT-4-0613.

The results on these benchmarks are shown across Table~\ref{table:likert-fc-gpt4} and Table~\ref{table:likert-vicuna-gpt4}.

\begin{table*}[t!]
\fontsize{8}{10}\selectfont
\centering
\caption{\footnotesize Pearson, Kendall-Tau, Spearman correlation with data generated by \textsc{GPT-4-0613}. All scores were sampled across 3 inferences. The best comparable statistics are \textbf{bolded} and second best \underline{underlined}.}
\resizebox{\textwidth}{!}{\begin{tabular}{lcccccc}
    \toprule
    \multicolumn{1}{c}{\multirow{3}{*}{\textbf{Evaluator LM}}}& \multicolumn{6}{c}{\textsc{Feedback Collection Test set (Generated by GPT-4-0613)}}\\
    \cmidrule(lr){2-7} & \multicolumn{3}{c}{\textsc{Seen Customized Rubrics}} & \multicolumn{3}{c}{\textsc{Unseen Customized Rubric}}\\ 
    \cmidrule(lr){2-4} \cmidrule(lr){5-7} & Pearson & Kendall-Tau & Spearman & Pearson & Kendall-Tau & Spearman\\ 
    \midrule
    \textsc{Llama2-Chat 7B}&0.485&0.422&0.478&0.463&0.402&0.465\\
    \textsc{Llama2-Chat 13B}&0.441&0.387&0.452&0.450&0.379&0.431\\
    \textsc{Llama2-Chat 70B}&0.572&0.491&0.564&0.558&0.477&0.549\\ 
    \textsc{Llama2-Chat 13B + Coarse.}&0.482&0.406&0.475&0.454&0.361&0.427\\ \midrule
    \textsc{Prometheus 7B}&\underline{0.860}&\textbf{0.781}&\textbf{0.863}&\underline{0.847}&\underline{0.767}&\underline{0.849}\\
    \textsc{Prometheus 13B }&\textbf{0.861}&\underline{0.776}&\underline{0.858}&\textbf{0.860}&\textbf{0.771}&\textbf{0.858}\\
    \textsc{GPT-3.5-Turbo-0613}&0.636&0.536&0.617&0.563&0.453&0.521\\
    \midrule
    \textsc{GPT-4-0314}&0.754&0.671&0.762&0.753&0.673&0.761\\
    \textsc{GPT-4-0613}&0.742&0.659&0.747&0.743&0.660&0.747\\
    \textsc{GPT-4 (Recent)}&0.745&0.659&0.748&0.733&0.641&0.728\\
    \bottomrule    
\end{tabular}}
\label{table:likert-fc-gpt4}
\end{table*}  

In Table~\ref{table:likert-fc-gpt4}, the performance of \textsc{Llama-2-Chat 13B} degrades over the 7B model and slightly improves when scaled up to 70B size, indicating that naively increasing the size of a model does not necessarily improve an LLM's evaluation capabilities. On the other hand, \textsc{Prometheus 13B} shows a +0.420 and +0.397 improvement over its base model \textsc{Llama2-Chat 13B} in terms of Pearson correlation on the seen and unseen rubric set, respectively. Moreover, it even outperforms \textsc{Llama2-Chat 70B}, \textsc{GPT-3.5-Turbo-0613}, and different versions of \textsc{GPT-4}. We conjecture the high performance of \textsc{Prometheus} is mainly because the instructions and responses within the test set might share a similar distribution with the train set we used (simulating a scenario where a user is interacting with a LLM) even if the score rubric holds unseen. Also, training on feedback derived from coarse-grained score rubrics (denoted as \textsc{Llama2-Chat 13B + Coarse}) only slightly improves performance, indicating the importance of training on a wide range of score rubric is important to handle customized rubrics that different LLM user or researcher would desire.

\begin{table*}[t!]
\fontsize{10}{16}\selectfont
\centering
\caption{\footnotesize Pearson, Kendall-Tau, Spearman correlation with scores sampled from \textsc{GPT-4-0613} across 3 inferences. Note that \textsc{GPT-4-0613} was sampled 6 times in total to measure self-consistency. The best comparable statistics are \textbf{bolded} and second best \underline{underlined} among baselines. We include GPT-4 as reference to show it self-consistency when inferenced multiple times.}
\resizebox{\textwidth}{!}{\begin{tabular}{lccccccccc}
    \toprule
    \multicolumn{1}{c}{\multirow{2}{*}{\textbf{Evaluator LM}}}& \multicolumn{3}{c}{\textsc{Vicuna Bench}} & \multicolumn{3}{c}{\textsc{MT Bench}} & \multicolumn{3}{c}{\textsc{FLASK EVAL}}\\ 
    \cmidrule(lr){2-4} \cmidrule(lr){5-7} \cmidrule(lr){8-10} & Pearson & Kendall-Tau & Spearman & Pearson & Kendall-Tau & Spearman & Pearson & Kendall-Tau & Spearman\\ 
    \midrule
    \textsc{Llama2-Chat 7B}&0.175&0.143&0.176&0.132 &0.113&0.143 & 0.271 & 0.180 & 0.235\\
    \textsc{Llama2-Chat 13B}&0.211&0.203&0.253&-0.020&-0.029 &-0.038 &0.265&0.182&0.235\\
    \textsc{Llama2-Chat 70B}&0.376&0.318&0.391&0.226&0.175&0.224&0.336&0.267&0.346\\
    \textsc{Llama2-Chat 13B + Coarse.}&0.307&0.196&0.245&\underline{0.417}&\underline{0.328} &\underline{0.420} &\textbf{0.517}&\textbf{0.349}&\underline{0.451}\\ \midrule
    \textsc{Prometheus-7B}&\underline{0.457}&\textbf{0.365}&\textbf{0.457}&0.293&0.216&0.295&0.367&0.285&0.371\\
    \textsc{Prometheus-13B}&\textbf{0.466}&\underline{0.346}&\underline{0.429}&\textbf{0.473}&\textbf{0.341}&\textbf{0.451}&\underline{0.467}&\underline{0.345}&\textbf{0.455}\\
    \textsc{GPT-3.5-Turbo-0613}&0.270&0.187&0.232&0.275&0.202&0.267&0.422&0.299&0.371\\
    \midrule
    \textsc{GPT-4-0314}&0.833&0.679&0.775&0.857&0.713&0.849&0.785&0.621&0.747\\
    \textsc{GPT-4-0613}&0.925&0.783&0.864&0.952&0.834&0.927&0.835&0.672&0.798\\
    \textsc{GPT-4 (RECENT)}&0.932&0.801&0.877&0.944&0.812&0.914&0.832&0.667&0.794\\
    \bottomrule    
\end{tabular}}
\label{table:likert-vicuna-gpt4}
\end{table*}

In Table~\ref{table:likert-vicuna-gpt4}, the trends of \textsc{Llama2-Chat} among different sizes hold similar; simply increasing size does not greatly improve the LLM's evaluation capabilities. On these benchmarks, \textsc{Prometheus} shows a +0.255, +0.493, and +0.202 improvement over its base model \textsc{Llama2-Chat-13B} in terms of Pearson correlation on the Vicuna Bench, MT Bench, and Flask Eval dataset, respectively. While \textsc{Prometheus} outperforms \textsc{Llama2-Chat 70B} and \textsc{GPT-3.5-Turbo-0613}, it lacks behind GPT-4. 
We conjecture that this might be because the instructions from the \textsc{Feedback Collection} and these evaluation datasets have different characteristics; the \textsc{Feedback Collection} are relatively long and detailed (e.g., I'm a city planner ... I'm looking for a novel and progressive solution to handle traffic congestion and air problems derived from population increase), while the datasets used for evaluation hold short (e.g., Can you explain about quantum mechanics?).

On the other hand, it is important to note that on the Flask Eval dataset, \textsc{Llama2-Chat 13B + Coarse} (specifically trained with the Flask Eval dataset) outperforms \textsc{Prometheus}. This indicates that training directly on the evaluation dataset might be the best option to acquire a task-specific evaluator LLM, and we further discuss this in Section~\ref{subsection:practioner_guide_eval}.

\begin{table*}[t!]
\centering
\fontsize{8}{10}\selectfont
\caption{\footnotesize Human Agreement accuracy among ranking datasets. The best comparable statistics are \textbf{bolded}.}
\resizebox{\textwidth}{!}{\begin{tabular}{lcccccc}
    \toprule
    \multicolumn{1}{c}{\multirow{2}{*}{\textbf{Evaluator LM}}}& \multicolumn{5}{c}{\textsc{HHH Alignment}} & \multicolumn{1}{c}{\textsc{MT Bench Human Judg.}}\\ 
    \cmidrule(lr){2-6} \cmidrule(lr){7-7} & Help. & Harm. & Hon. & Other & Total Avg. & Human Preference\\ 
    \midrule
    \textsc{Random}&50.00&50.00&50.00&50.00&50.00&34.26\\
    \textsc{StanfordNLP Reward Model}&69.49&60.34&52.46&51.16&58.82&44.79\\ 
    \textsc{ALMOST Reward Model}&74.58&67.24&\textbf{78.69}&86.05&76.02&49.90\\ 
    \textsc{Llama2-Chat 7B}&66.10&81.03&70.49&74.42&72.85&51.78\\
    \textsc{Llama2-Chat 13B}&74.58&87.93&55.74&79.07&73.76&52.34\\
    \textsc{Llama2-Chat 70B}&66.10&\textbf{89.66}&67.21&74.42&74.21&53.67\\
    \textsc{Llama2-Chat 13B + Coarse.}&68.74&68.97&65.57&67.44&67.42&46.89\\ \midrule
    \textsc{GPT-3.5-Turbo-0613}&76.27&87.93&67.21&86.05&78.73&57.12\\
    \textsc{Prometheus 7B}&69.49&84.48&\textbf{78.69}&\textbf{90.70}&\textbf{80.09}&55.14\\
    \textsc{Prometheus 13B}&\textbf{81.36}&82.76&75.41&76.74&79.19&\textbf{57.72}\\
    \midrule
    \textsc{GPT-4-0613}&91.53&93.10&85.25&83.72&88.69&63.87\\
    \bottomrule    
\end{tabular}}
\label{table:human-correlation}
\end{table*}  

\subsection{Can \textsc{Prometheus} Function as a Reward Model?}\label{subsec:5.3}

We conduct experiments on 2 human preference datasets: HHH Alignment~\citep{askell2021general} and MT Bench Human Judgment~\citep{zheng2023judging} that use a ranking grading scheme. In Table~\ref{table:human-correlation}, results show that prompting \textsc{Llama-2-Chat} surprisingly obtains reasonable performance, which we conjecture might be the effect of using a base model that is trained with Reinforcement Learning from Human Feedback (RLHF). When training on feedback derived from coarse-grained score rubrics (denoted as \textsc{Llama2-Chat 13B + Coarse}), it only hurts performance. On the other hand, \textsc{Prometheus 13B} shows a +5.43\% and +5.38\% margin over its base model \textsc{Llama2-Chat-13B} on the HHH Alignment and MT Bench Human Judgement dataset, respectively. These results are surprising because they indicate that training on an absolute grading scheme could also improve performance on a ranking grading scheme even without directly training on ranking evaluation instances. Moreover, it shows the possibilities of using a generative LLM (\textsc{Prometheus}) as a reward model for RLHF~\citep{kim2023aligning}. We leave the exploration of this research to future work.

\section{Discussions and Analysis}


\subsection{Why is it important to include Reference Materials?}
Evaluating a given response without any reference material is a very challenging task (i.e., Directly asking to decide a score only when an instruction and response are given), since the evaluation LM should be able to (1) know what the important aspects tailored with the instruction is, (2) internally estimate what the answer of the instruction might be, and (3) assess the quality of responses based on the information derived from the previous two steps. Our intuition is that by incorporating each component within the reference material, the evaluator LM could solely focus on assessing the quality of the response instead of determining the important aspects or solving the instruction. Specifically, we analyze the role of each component as follows:
\begin{itemize}
    \item \textbf{Score Rubric}: Giving information of the the pivotal aspects essential for addressing the given instruction. Without the score rubric, the evaluator LM should inherently know what details should be considered from the given instruction.
    \item \textbf{Reference Answer}: Decomposing the process of estimating a reference answer and evaluating it at the same time into two steps. Since the reference answer is given as an additional input, the evaluator LM could only focus on evaluating the given response. This enables to bypass a natural proposition that if an evaluator LM doesn't have the ability to solve the problem, it's likely that it cannot evaluate different responses effectively as well.
\end{itemize}

As shown in Table~\ref{table:ablation}, we conduct an ablation experiment by excluding each reference material and also training only on the score rubric without generating a feedback. Additionally, we also ablate the effect of using different model variants (Llama-2, Vicuna, Code-Llama) instead of Llama-2-Chat.

\paragraph{Training Ablation} The results indicate that each component contributes orthogonally to \textsc{Prometheus}'s superior evaluation performance. Especially, excluding the reference answer shows the most significant amount of performance degradation, supporting our claim that including a reference answer relieves the need for the evaluator LM to internally solve the instruction and only focus on assessing the response. Also, while excluding the score rubric on the \textsc{Feedback Bench} does not harm performance a lot, the performance drops a lot when evaluating on Vicuna Bench. As in our hypothesis, we conjecture that in order to generalize on other datasets, the role of providing what aspect to evaluate holds relatively crucial.

\paragraph{Model Ablation} To test the effect using \textsc{Llama2-Chat}, a model that has been instruction-tuned with both supervised fine-tuning and RLHF, we ablate by using different models as a starting point. Results show that different model choices do not harm performance significantly, yet a model trained with both supervised fine-tuning and RLHF shows the best performance, possibly due to additional training to follow instructions. However, we find that using Code-Llama has some benefits when evaluating on code domain, and we discuss the effect on Section~\ref{subsection:practioner_guide_train}.

\begin{table*}[t!]
\centering
\caption{\footnotesize Pearson, Kendall-Tau, Spearman correlation with data generated by \textsc{GPT-4-0613} (Feedback Collection Test set) and scores sampled from \textsc{GPT-4-0613} across 3 inferences (Vicuna Bench).}
\fontsize{5}{6}\selectfont
\resizebox{\textwidth}{!}{\begin{tabular}{lcccc}
    \toprule
    \multicolumn{1}{c}{\multirow{3}{*}{\textbf{Evaluator LM}}}& \multicolumn{2}{c}{\textsc{Feedback Collection Test set}} & \textsc{Vicuna Bench}\\
    \cmidrule(lr){2-4} & Seen Score Rubric & Unseen Score Rubric & -\\ 
    \cmidrule(lr){2-2} \cmidrule(lr){3-3} \cmidrule(lr){4-4} & Pearson & Pearson & Pearson\\ 
    \midrule
    \textsc{Prometheus 7B}&0.860&0.847&0.457\\
    \midrule
    \multicolumn{4}{c}{\textbf{Training Ablation}}\\
    \midrule
    \textsc{w/o Score Rubric}&0.837&0.745&0.355\\
    \textsc{w/o Feedback Distillation}&0.668&0.673&0.413\\
    \textsc{w/o Reference Answer}&0.642&0.626&0.349\\
    \midrule
    \multicolumn{4}{c}{\textbf{Model Ablation}}\\
    \midrule
    \textsc{Llama-2 7B Baseline}&0.839&0.818&0.404\\
    \textsc{Vicuna-v1.5 7B Baseline}&0.860&0.829&0.430\\
    \textsc{Code-Llama 7B Baseline}&0.823&0.761&0.470\\
    \bottomrule    
\end{tabular}}
\label{table:ablation}
\end{table*}

\subsection{Narrowing Performance Gap to GPT-4 Evaluation}
The observed outcomes, in which \textsc{Prometheus} consistently surpasses GPT-4 based on human evaluations encompassing both scores and quality of feedback, as well as correlations in the \textsc{Feedback Bench}, are indeed noteworthy. We firmly posit that these findings are not merely serendipitous and offer the following justifications:
\begin{itemize}
    \item Regarding results on \textsc{Feedback Bench}, our model is directly fine-tuned on this data, so it's natural to beat GPT-4 on a similar distribution test set if it is well-trained. In addition, for GPT-4, we compare the outputs of \textbf{inferencing} on the instructions and \textbf{augmenting} new instances, causing the self-consistency to be lower.
    \item Regarding score correlation for human evaluation, our model shows similar or slightly higher trends. First, our human evaluation set-up excluded all coding or math-related questions, which is where it is non-trivial to beat GPT-4 yet. Secondly, there's always the margin of human error that needs to be accounted for. Nonetheless, we highlight that we are the first work to argue that an open-source evaluator LM could closely reach GPT-4 evaluation \textit{only when the appropriate reference materials are accompanied}.
    \item As shown in Figure~\ref{figure:human_eval_why_feedback_gpt4_prometheus}, \textsc{Prometheus} tends to be critical compared to GPT-4. We conjecture this is because since it is specialized for evaluation, it acquires the characteristics of seeking for improvement when assessing responses.
\end{itemize}

\subsection{Qualitative Examples of Feedback Generated by \textsc{Prometheus}}
We present five qualitative examples to compare the feedback generated by \textsc{Prometheus} and GPT-4 in Appendix~\ref{appendix:qualitative_example}. Specifically, Figure~\ref{figure:too_abstract} shows an example where human annotators labeled that GPT-4 generate an abstract/general feedback not suitable for criticizing the response. Figure~\ref{figure:over_critical} shows an example where human annotators labeled that \textsc{Prometheus} generate overly critical feedback. Figure~\ref{figure:tie} shows an example of human annotators labeled as a tie. In general, \textsc{Prometheus} generates a detailed feedback criticizing \textit{which} component within the response is wrong and seek improvement. This qualitatively shows that \textsc{Prometheus} could function as an evaluator LM.

Moreover, we present an example of evaluating python code responses using \textsc{Prometheus}, GPT-4, and Code-Llama in Figure~\ref{figure:code2}. We discuss the effect of using a base model specialized on code domain for code evaluation in Section~\ref{subsection:practioner_guide_train}.

\subsection{A Practitioner's Guide for Directly Using \textsc{Prometheus} Evaluation}\label{subsection:practioner_guide_eval}
\paragraph{Preparing an Evaluation Dataset} As shown in the previous sections, \textsc{Prometheus} functions as a good evaluator LM not only on the \textsc{Feedback Bench} (a dataset that has a similar distribution with the dataset it was trained on), but also on other widely used evaluation datasets such as the Vicuna Bench, MT Bench, and Flask Eval. As shown in Figure~\ref{figure:main_figure}, users should prepare the instruction dataset they wish to evaluate their target LLM on. This could either be a widely used instruction dataset or a custom evaluation users might have.

\paragraph{Deciding a Score Rubric to Evaluate on} The next step is to choose the score rubric users would want to test their target LLM on. This could be confined to generic metrics such as helpfulness/harmlessness, but \textsc{Prometheus} also supports fine-grained score rubrics such as ``Child-Safety'', ``Creativity'' or even ``Is the response formal enough to send to my boss''.

\paragraph{Preparing Reference Answers} While evaluating without any reference is also possible, as shown in Table~\ref{table:ablation}, \textsc{Prometheus} shows superior performance when the reference answer is provided. Therefore, users should prepare the reference answer they might consider most appropriate based on the instructions and score rubrics they would want to test on. While this might require additional cost to prepare, there is a clear trade-off in order to improve the precision or accuracy of the overall evaluation process, hence it holds crucial.

\paragraph{Generating Responses using the Target LLM} The last step is to prepare responses acquired from the target LLM that users might want to evaluate. By providing the reference materials (score rubrics, reference answers) along with the instruction and responses to evaluate on, \textsc{Prometheus} generates a feedback and a score. Users could use the score to determine how their target LLM is performing based on customized criteria and also refer to the feedback to analyze and check the properties and behaviors of their target LLM. For instance, \textsc{Prometheus} could be used as a good alternative for GPT-4 evaluation while training a new LLM. Specifically, the field has not yet come up with a formalized procedure to decide the details of instruction-tuning or RLHF while developing a new LLM. This includes deciding how many training instances to use, how to systematically decide the training hyperparameters, and quantitatively analyzing the behaviors of LLMs across multiple versions. Most importantly, users might not want to send the outputs generated by their LLMs to OpenAI API calls. In this regard, \textsc{Prometheus} provides an appealing solution of having control over the whole evaluation process, also supporting customized score rubrics.

\subsection{A Practitioner's Guide for Training a New Evaluation Model}\label{subsection:practioner_guide_train}
Users might also want to train their customized evaluator LM as \textsc{Prometheus} for different use cases. As shown in Table~\ref{table:likert-vicuna-gpt4}, training directly on the Flask dataset (denoted as \textsc{Llama2-Chat 13B + Coarse}) shows a higher correlation with GPT-4 on the Flask Eval dataset compared to \textsc{Prometheus} that is trained on the \textsc{Feedback Collection}. This implies that directly training on a target feedback dataset holds the best performance when evaluating on it. Yet, this requires going through the process of preparing a new feedback dataset (described in Section~\ref{subsec:data_collection}). This implies that there is a trade-off between obtaining a strong evaluator LM on a target task and paying the initial cost to prepare a new feedback dataset. In this subsection, we provide some guidelines for how users could also train their evaluator LM using feedback datasets.

\paragraph{Preparing a Feedback Dataset to train on} As described in Section~\ref{section:3}, some important considerations to prepare a new feedback dataset are: (1) including as many reference materials as possible, (2) maintaining a uniform length among the reference answers for each score (1 to 5) to prevent undesired length bias, (3) maintaining a uniform score distribution to prevent undesired decision bias. While we did not explore the effect of including other possible reference materials such as a ``Score 1 Reference Answer'' or ``Background Knowledge'' due to limited context length, future work could also explore this aspect. The main intuition is that providing more reference materials could enable the evaluator LM to solely focus on evaluation instead of solving the instruction.

\paragraph{Choosing a Base Model to Train an Evaluator LM} As shown in Figure~\ref{figure:code2}, we find that training on \textsc{Code-Llama} provides more detailed feedback and a reasonable score decision when evaluating responses on code domains (7 instances included within the Vicuna Bench dataset). This indicates that choosing a different base model based on the domain to evaluate might be crucial when designing an evaluator LM. We also leave the exploration of training an evaluator LM specialized on different domains (e.g., code and math) as future work.






\section{Conclusion}
In this paper, we discuss the possibility of obtaining an open-source LM that is specialized for fine-grained evaluation. While text evaluation is an inherently difficult task that requires multi-faceted considerations, we show that by incorporating the appropriate reference material, we can effectively induce evaluation capability into an LM. We propose a new dataset called the \textsc{Feedback Collection} that encompasses thousands of customized score rubrics and train an open-source evaluator model, \textsc{Prometheus}. Surprisingly, when comparing the correlation with human evaluators, \textsc{Prometheus} obtains a Pearson correlation on par with GPT-4, while the quality of the feedback was preferred over GPT-4 58.62\% of the time. When comparing Pearson correlation with GPT-4, \textsc{Prometheus} shows the highest correlation even outperforming GPT-3.5-Turbo. Lastly, we show that \textsc{Prometheus} shows superior performance on human preference datasets, indicating its possibility as an universal reward model. We hope that our work could stimulate future work on using open-source LLMs as evaluators instead of \textit{solely} relying on proprietary LLMs.


\subsubsection*{Acknowledgments}
This work was partly supported by KAIST-NAVER Hypercreative AI Center and Institute of Information \& communications Technology Planning \& Evaluation (IITP) grant funded by the Korea government (MSIT) (No.2022-0-00264, Comprehensive Video Understanding and Generation with Knowledge-based Deep Logic Neural Network, 40\%; No.2021-0-02068, Artificial Intelligence Innovation Hub, 20\%). We thank Minkyeong Moon, Geonwoo Kim, Minkyeong Cho, Yerim Kim, Sora Lee, Seunghwan Lim, Jinheon Lee, Minji Kim, and Hyorin Lee for helping with the human evaluation experiments. We thank Se June Joo, Dongkeun Yoon, Doyoung Kim, Seonghyeon Ye, Gichang Lee, and Yehbin Lee for helpful feedback and discussions.
\clearpage


\bibliographystyle{iclr2024_conference}

\clearpage
\appendix

\begin{figure*}[t!]
\centering
    \includegraphics[width=0.8\linewidth]{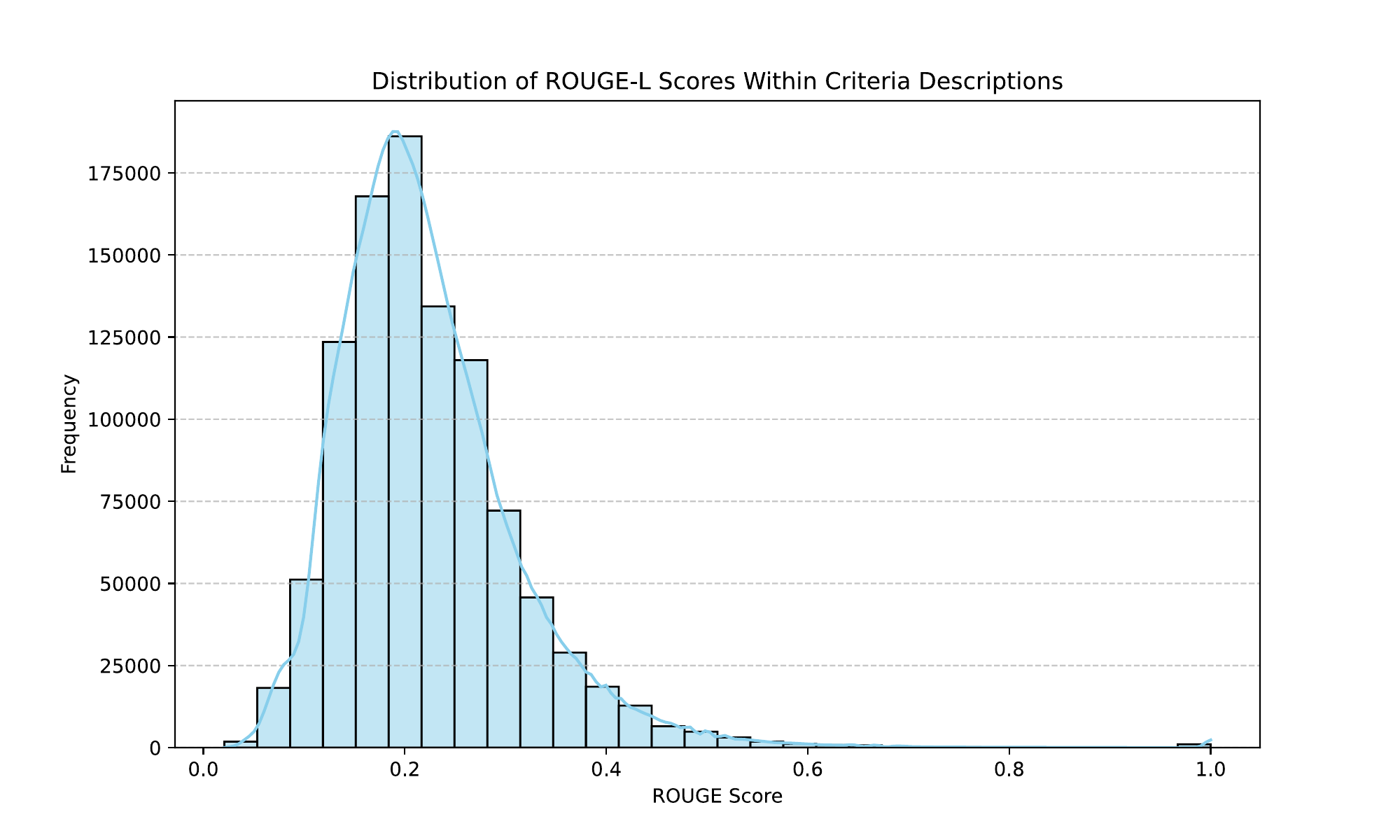}
    \caption{Rouge-L score distribution among two randomly sampled score rubrics from the \textsc{Feedback Collection}. A left-skewed distribution with low values shows the score rubrics are diverse.}
    \label{figure:rouge_train}
\end{figure*}

\begin{figure*}[t!]
\centering
    \includegraphics[width=0.7\linewidth]{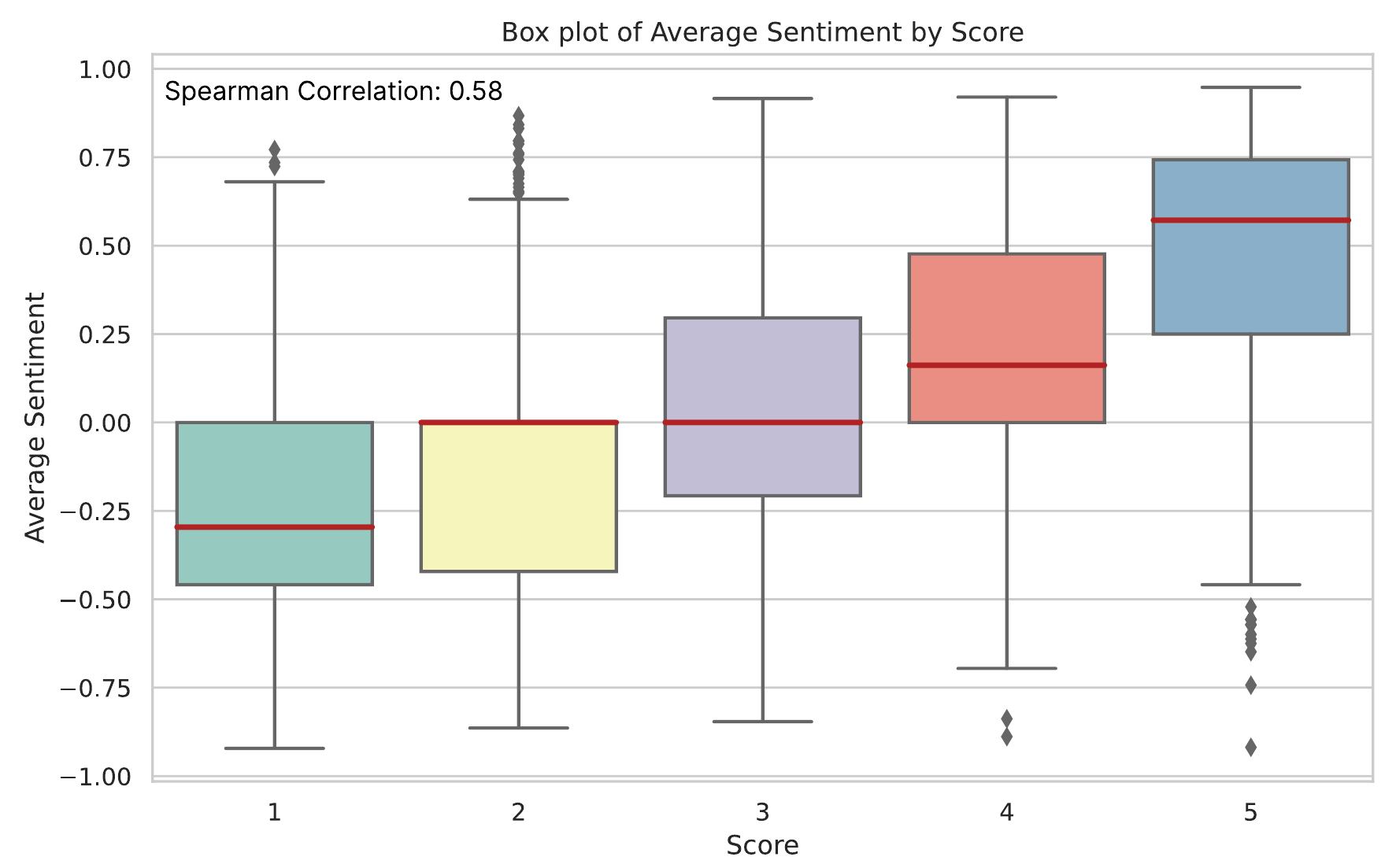}
    \caption{Box and whisker plot for average sentiment per score description. A linearly increasing trend is crucial for the evaluator LM to decide a score in an Absolute Scoring setting.}
    \label{figure:sentiment}
\end{figure*}

\section{Analysis of the \textsc{Feedback Collection} Dataset}\label{appendix:feedback_collection}
In this section, we provide a comprehensive analysis of the characteristics of the \textsc{Feedback Collection} dataset. To ensure the quality, we answer each question one by one, emphasizing our main considerations during the creation of the dataset.

\paragraph{Are the Score Criteria Diverse Enough?} Following previous work~\citep{wang2022self,honovich2022unnatural}, we plot the rouge-L distribution between two instances among our whole set of 1K score rubrics. Specifically, we use the score criteria (description of the criteria) and measure the rouge-L value between the two score criteria. Figure~\ref{figure:rouge_train} shows the overall distribution plot. The results indicate that each criteria does not overlap with one another, ensuring that we include many \textit{novel} score rubrics in our training set.

\begin{figure*}[t!]
\centering
    \includegraphics[width=0.9\linewidth]{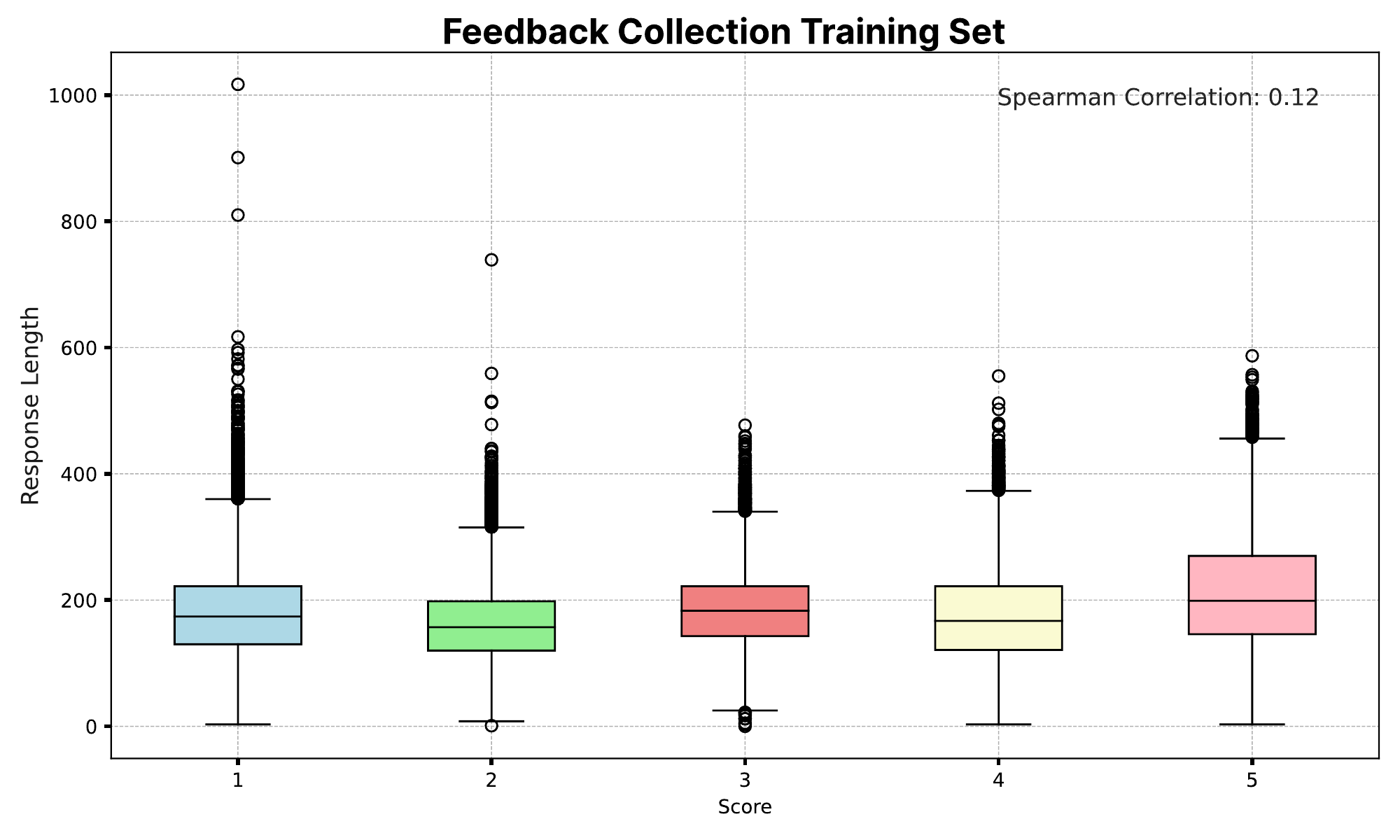}
    \caption{Box and whisker plot plotting the length distribution of responses on each score range. We check whether there is a length bias (i.e., a higher score given for longer responses).}
    \label{figure:length_bias}
\end{figure*}
\begin{table*}[t!]
\centering
\caption{Distinct N-gram measured on each component of the training instance. A higher diversity ratio indicates that each component tends to be more diverse.}
\begin{tabular}{@{}ccc@{}}
\toprule
Distinct N-gram         & Bigram Diversity Ratio & Trigram Diversity Ratio \\ \midrule
Instruction   & 0.43                   & 0.79                    \\
Reference     & 0.43                   & 0.82                    \\
Score Rubric & 0.60                   & 0.81                    \\ \midrule
Responses     & 0.32                   & 0.77                    \\
Feedback      & 0.26                   & 0.66                    \\ \bottomrule
\end{tabular}
\label{tab:distinc_ngram}
\end{table*}

\paragraph{Are the Score Descriptions Well Formulated?} Another component in the score rubric is a description of each score (i.e., A comprehensive reason why a score of $i$ $(1 \leq i \leq 5$ should be given). In an Absolute Scoring setting, it is important to evaluate the given response based on the score descriptions instead of giving a score of 1 for all responses that lack a minor detail or giving a score of 5 for all responses that seem to be good on the surface. Due to these reasons, the role of the score descriptions hold crucial, where the main role is to show a monotonically increasing tendency of sentiment, not dramatically. Figure~\ref{figure:sentiment} shows that the \textsc{Feedback Collection} holds a smoothly increasing sentiment tendency for each score description. This ensures the quality of the score rubric, confirming that it plays a role in deciding the score.

\paragraph{Is there a length bias among the Responses?} Previous work has demonstrated that when LMs are used as evaluators, they tend to give higher scores to longer responses~\citep{alpaca_eval,dubois2023alpacafarm,zheng2023judging}. In order to minimize this effect during fine-tuning \textsc{Prometheus}, one of our main consideration was to maintain a length distribution equal among the score range of 1 to 5. As shown in Figure~\ref{figure:length_bias}, most of the responses within the \textsc{Feedback Collection} maintained a similar length among different scores (near 200 tokens). We also include a comprehensive analysis of whether \textsc{Prometheus} possessed any length bias during evaluation in Appendix~\ref{section:length_bias_evaluation}.

\paragraph{Are the Instructions, Responses, and Feedback Diverse as Well?} In addition to the analysis of the score rubric and responses, we also analyze whether the instructions, responses, and feedback within the \textsc{Feedback Collection} are diverse enough. For this purpose, we examine the bigram and trigram ratios. The results are shown in Table~\ref{tab:distinc_ngram}, indicating a variety in how terms are expressed, and our findings suggest a moderate level of diversity. While there is some term repetition, the dataset also showcases a notable range of expressions.

\section{Analysis of the \textsc{Feedback Bench} Evaluation Dataset}\label{appendix:feedback_bench}

In this section, we provide a analysis of whether the \textsc{Feedback Bench} consists of unseen score rubrics against the score rubrics from the \textsc{Feedback Collection}.

\begin{figure*}[t!]
\centering
    \includegraphics[width=0.8\linewidth]{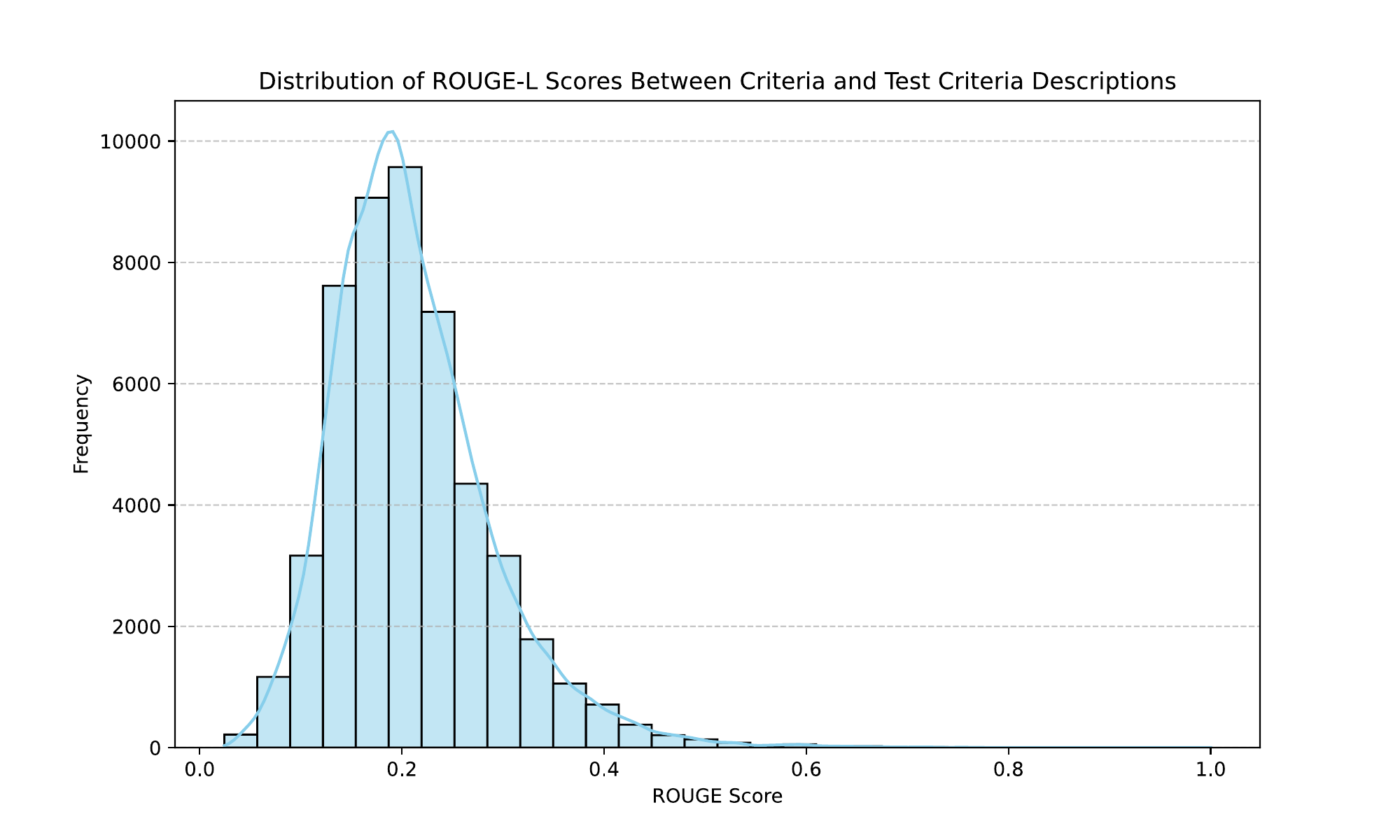}
    \caption{Rouge-L score distribution among a randomly sampled score rubric from the \textsc{Feedback Collection} and a score rubric from the \textsc{Feedback Bench}. A left-skewed distribution with low values shows that they do not overlap with each other, hence meaning that a \textsc{unseen} score rubric assumption is satisfied.}
    \label{figure:rouge_test}
\end{figure*}

\paragraph{Does the testset maintain Unseen Score Rubrics?} One of the main considerations of our experiments in Section~\ref{subsec:5.2} using the \textsc{Feedback Bench} was testing whether \textsc{Prometheus} could generalize to unseen \textit{customized} score rubrics. For this purpose, we built an unseen customized rubric subset. We plot the rouge-L distribution between a random score rubric within the \textsc{Feedback Collection} and a random score rubric within the \textsc{Feedback Bench}. As shown in Figure~\ref{figure:rouge_test}, there is a low overlap among the train and test sets, confirming that the \textsc{Feedback Bench} is valid to be claimed as an \textit{unseen} test set to measure the evaluation capability of evaluator LMs.

\section{Fine-tuning and Inference Details of \textsc{Prometheus}}\label{appendix:hyperparameter_details}

We use 8xA100 (80GB) GPUs to train our models with PyTorch Fully-Sharded Data Parallel (FSDP) option. The code we used for training and inference is the official Llama2 fine-tuning code released by Meta AI\footnote{\url{https://github.com/facebookresearch/llama-recipes}}.
The hyper-parameters we used are the basic settings in the fine-tuning code except for the training batch size which was set according to the model size: for 7B models we used 28 and for 13B models we used 20 to fully leverage GPU memory. Note that in the official Llama2 fine-tuning code, the loss is only calculated on the feedback and score decision, not the instruction. We empirically find that not masking out the instruction leads to poor performance while evaluating responses.
The detailed hyper-parameters are shown in Table~\ref{tab:config_train}.

\begin{table*}[t!]
\fontsize{6}{8}\selectfont
\centering
\caption{Hyperparameters used for fine-tuning \textsc{Prometheus}.}
\begin{tabular}{ccccccc}
\toprule
Model & Base Model & Batch size & LR & LR Scheduler & Optimizer & Max Length (Input \& Output)\\
\midrule
\textsc{Prometheus}-7B  & Llama-2-Chat-7B & 28 & 1e-5 & StepLR & AdamW   & 4096  \\
\textsc{Prometheus}-13B & Llama-2-Chat-13B & 20 & 1e-5 & StepLR &AdamW  & 4096\\
\bottomrule
\multicolumn{1}{l}{} & \multicolumn{1}{l}{} & \multicolumn{1}{l}{} & \multicolumn{1}{l}{} & \multicolumn{1}{l}{}
\end{tabular}%
\label{tab:config_train}
\end{table*}

\begin{table*}[t!]
\fontsize{8}{10}\selectfont
\centering
\caption{Hyperparameters used for inferencing \textsc{Prometheus}, GPT-3.5-Turbo, and GPT-4. Verbalizer denotes accepting outputs such as ''[Score 5]'' or ''Score: 4 out of 5'' whereas the exact format is ''[Result] 5'' (format is mentioned concretely within the instruction given to the evaluator LM). Even after applying a verbalizer, Llama-2-Chat is not able to generate a score decision that could easily be parsed, highlighting the benefits of fine-tuning it on feedback data.}
\begin{tabular}{ccccccc}
\toprule
Params & Model & Temperature & Top-p & Repetition Penalty   & Max Output Length & Verbalizer\\
\midrule
7B & Llama-2-Chat-7B & 1.0 & 0.9 & 1.03 & 256 & Yes\\
13B & Llama-2-Chat-13B & 1.0 & 0.9 & 1.03 & 256 & Yes\\
70B & Llama-2-Chat-70B & 1.0 & 0.9 & 1.03 & 256 & Yes\\
7B     & \textsc{Prometheus}-7B  & 1.0 & 0.9 & 1.03         & 256  & No\\
13B    & \textsc{Prometheus}-13B & 1.0 & 0.9 & 1.03          & 256 & No\\
-    & GPT-3.5-Turbo & 1.0 & 0.9 & -          & 256 & No\\
-    & GPT-4 & 1.0 & 0.9 & -          & 256  & No\\
\bottomrule
\multicolumn{1}{l}{} & \multicolumn{1}{l}{} & \multicolumn{1}{l}{} & \multicolumn{1}{l}{} & \multicolumn{1}{l}{}
\end{tabular}%
\label{tab:config_eval}
\end{table*}

For inference, we use the hyper-parameters as shown in Table~\ref{tab:config_eval}. When inferencing with the naive Llama-2-Chat model (not trained on the \textsc{Feedback Collection}) it was extremely difficult to steer the model to generate a final score in the form to be easily parsed (e.g., ``[RESULT] 3''). While in-context learning (ICL) could solve this issue, most of our instances contained a maximum of 3072 tokens, so we could not utilize demonstrations during inference. Therefore, we empirically found patterns such as ``[SCORE 5]'' or ``Score: 4 out of 5'' and applied verbalizer to map those outputs to a final score decision. This highlights the benefits of directly training to generate in a structured format as \textsc{Prometheus}. On the other hand, we also find that proprietary LLMs such as GPT-3.5-Turbo and GPT-4 excel at generating structured outputs when the prompt is adeptly given. Also, note that we found that if we set the temperature to 0.0, evaluator LMs are not able to generate meaningful feedback compared to using a temperature of 1.0.

\section{Training a Evaluator LM on Coarse-grained Score Rubrics}\label{appendix:coarse_grained}

For the purpose of exploring the benefits of training on thousands of fine-grained and customized score rubrics, we employ a baseline of only training on relatively \textit{coarse-grained} score rubrics. Since the \textsc{Feedback Collection}'s instructions are closely tied with the score rubrics during its creation process, we could not directly use it and only change the score rubrics into coarse-grained ones. 

So, we used the Flask dataset~\citep{ye2023flask} and split it into training data and evaluation data. The evaluation data is denoted as Flask Eval throughout the paper. Specifically, the Flask dataset consists of 1.7K instructions acquired across conventional NLP datasets and instruction datasets. Also, there exists 76.5K responses acquired across 15 response LMs. Each instance has a score rubric among 12 options (Logical Robustness, Logical Correctness, Logical Efficiency, Factuality, Commonsense Understanding, Harmlessness, Readability, Comprehension, Insightfulness, Completeness, Metacognition, Conciseness). While these 12 score rubrics are more fine-grained and diverse compared to previous works only using helpfulness and harmlessness, they are coarse-grained compared to the thousands of score rubrics included within the \textsc{Feedback Collection}, so we denote as coarse-grained in this work. 

Among the 1.5K instructions \& 67.5K responses as training data, we found that the score distribution is extremely skewed towards the score of 5. We distributed the instances so that the number of instances within the score range of 1 to 5 remains equal, which leads to 30K training instances. We trained the Llama-2-Chat model on the Flask train set, which led to one of our baselines denoted as \textsc{Llama-2-Chat + Coarse} in Table~\ref{table:likert-fc-gpt4}, Table~\ref{table:likert-vicuna-gpt4}, Table~\ref{table:human-correlation}, Table~\ref{table:ablation}.

\begin{figure*}[t!]
\centering
    \includegraphics[width=0.9\linewidth]{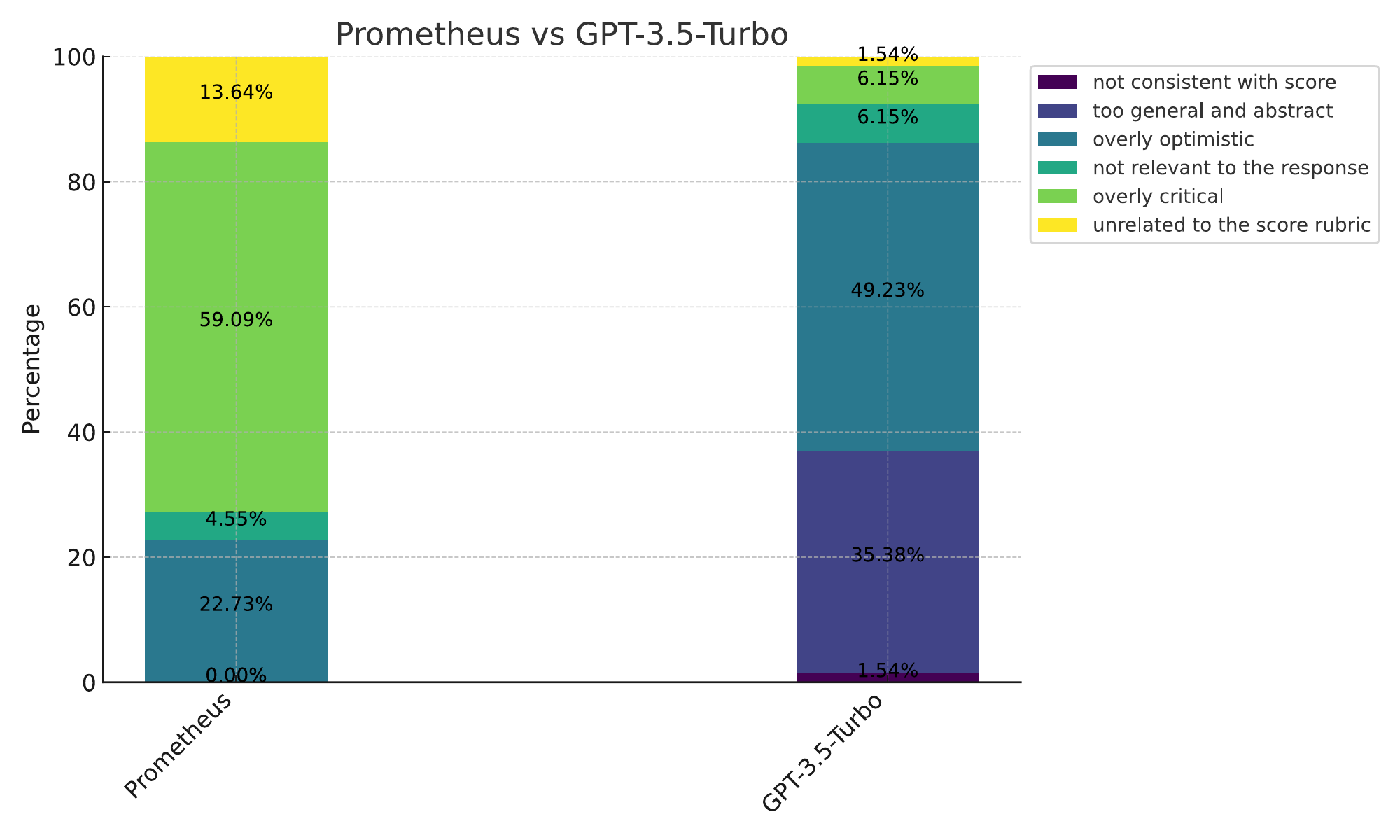}
    \caption{The reason why GPT-3.5-Turbo's or Prometheus's feedback was not chosen over the other. \textsc{Prometheus} generates less abstract and general feedback, but tends to write overly critical ones.}
    \label{figure:human_eval_why_feedback_prometheus_chatgpt}
\end{figure*}

\begin{figure*}[t!]
\centering
    \includegraphics[width=0.9\linewidth]{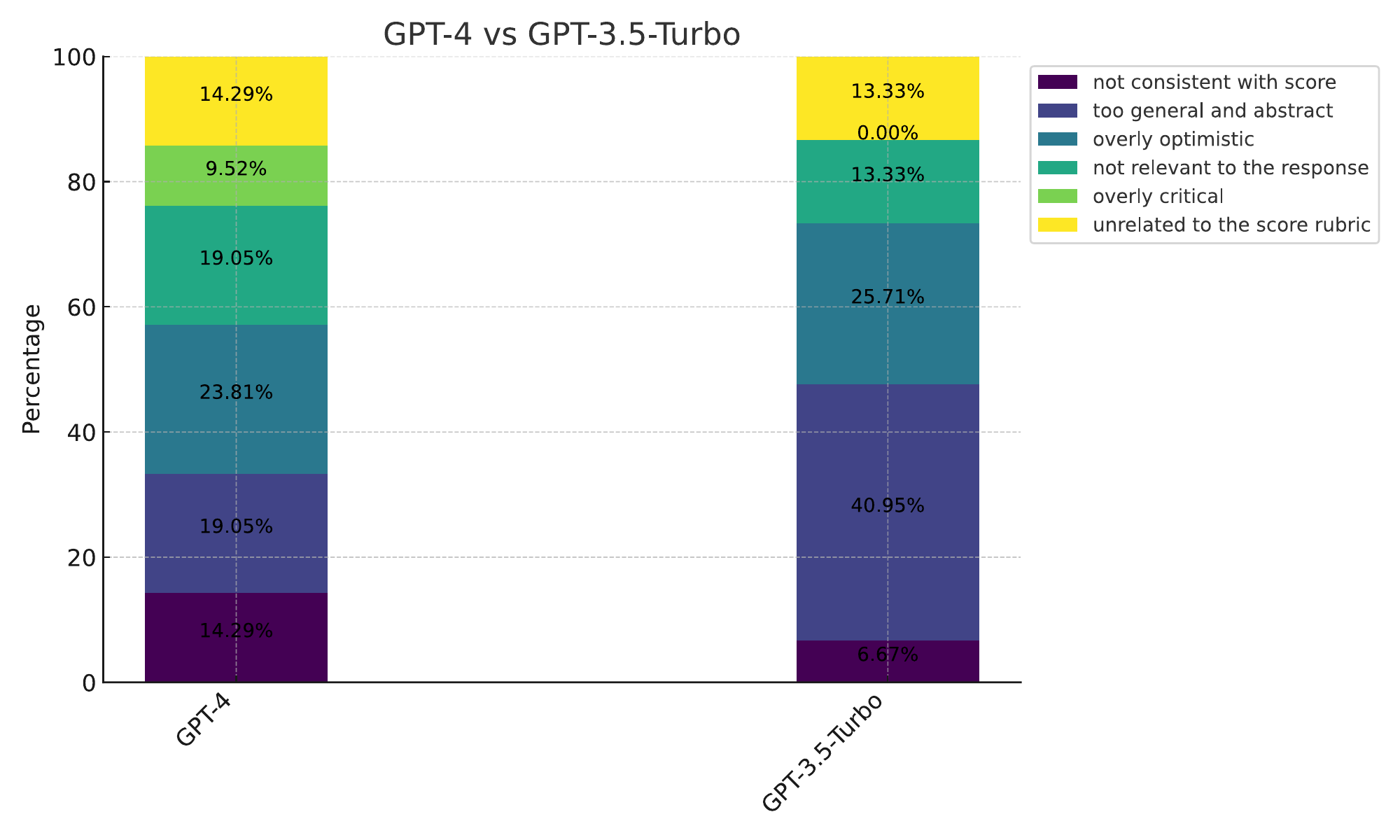}
    \caption{The reason why GPT-4's or GPT-3.5-Turbo's feedback was not chosen over the other. GPT-4 generates less abstract and general feedback, but tends to write overly critical ones.}
    \label{figure:human_eval_why_feedback_chatgpt_gpt4}
\end{figure*}

\section{Pairwise Comparison of the Quality of the Feedback}\label{appendix:comparison_experiment}
In this section, we further explain the experimental setting and present additional results \& analysis for the experiment of comparing the quality of the generated feedback (Section~\ref{subsec:5.1}). 

In addition to Figure~\ref{figure:human_eval_why_feedback_gpt4_prometheus}, the reason \textit{why} each annotator rejected the feedback from either \textsc{Prometheus}, GPT-3.5-Turbo, GPT-4 is shown in Figure~\ref{figure:human_eval_why_feedback_prometheus_chatgpt} and Figure~\ref{figure:human_eval_why_feedback_chatgpt_gpt4}.

The results further support our claim that \textsc{Prometheus} tends to be critical over GPT-4 and GPT-3.5-Turbo. Interestingly, GPT-4 was considered to be more critical compared to GPT-3.5-Turbo and the gap was even wider when comparing GPT-3.5-Turbo and \textsc{Prometheus}. This indicates that \textsc{Prometheus} can serve as a critical judge when evaluating responses generated by LLMs, but it could also be biased towards not being optimistic generally. The degree of being critical could be useful or a limitation based on different use cases. For instance, we conjecture that it could be helpful when analyzing the limitations of LLMs or providing feedback as supervision to further improve a target LLM (e.g., RLHF), yet we leave this exploration to future work.

\section{Is there a Length Bias during Evaluation?}\label{section:length_bias_evaluation}

\begin{figure*}[t!]
\centering
    \includegraphics[width=0.8\linewidth]{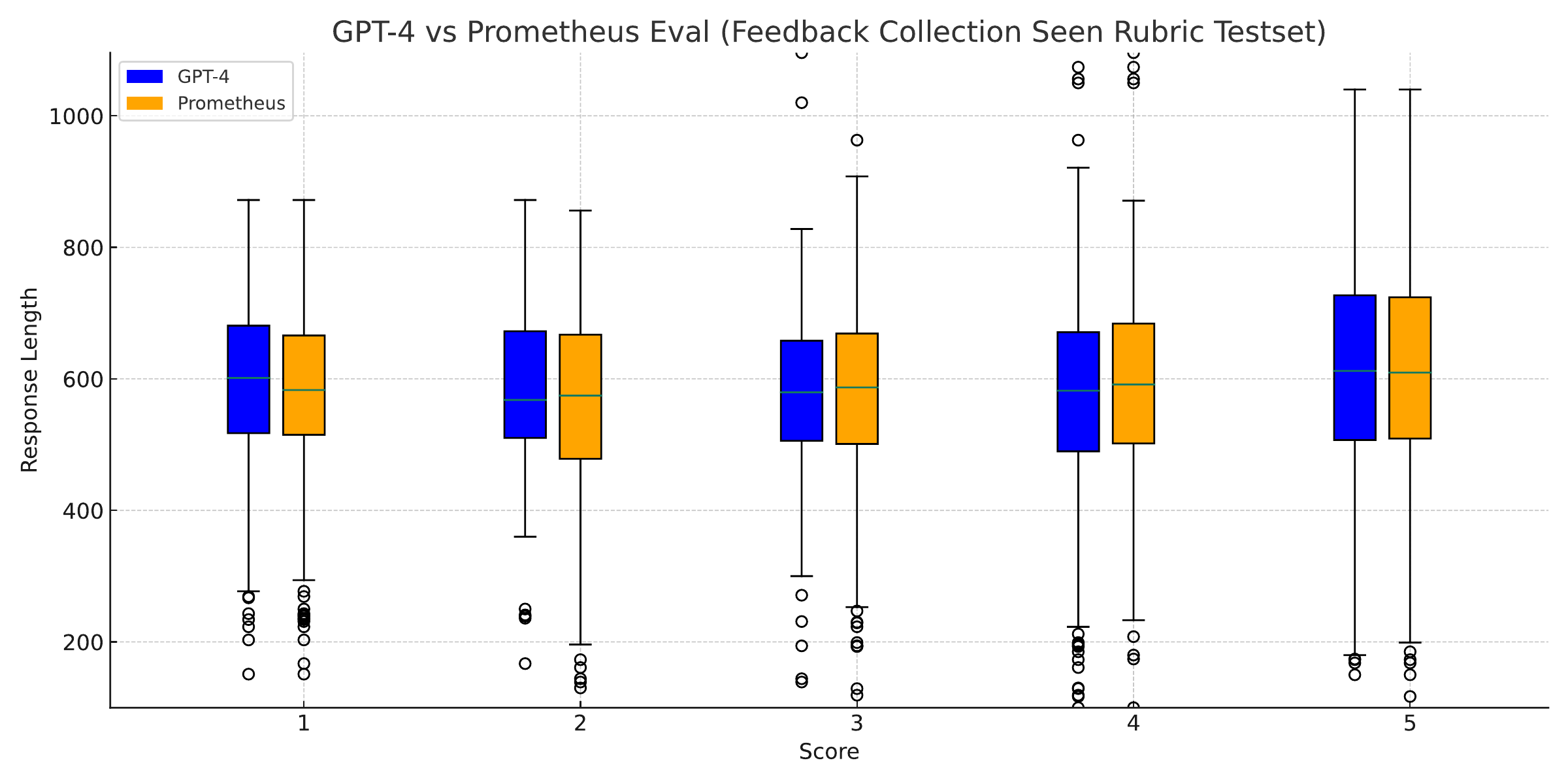}
    \caption{Box and whisker plot describing a relationship between a given response and its corresponding score. We check if the response lengths correlate with its scores.}
    \label{figure:length-bias-fc-seen-combined}
\end{figure*}

\begin{figure*}[t!]
\centering
    \includegraphics[width=0.8\linewidth]{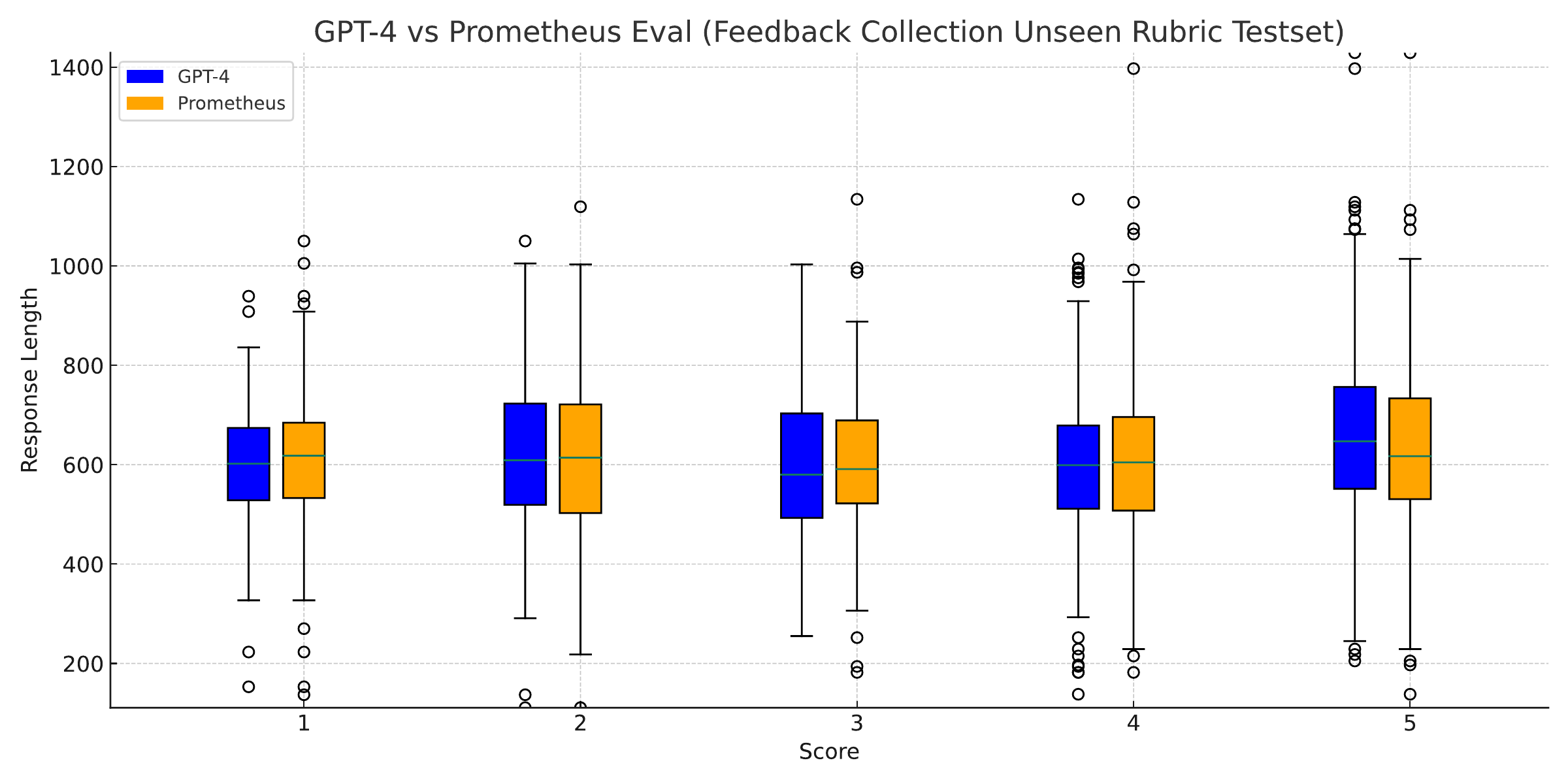}
    \caption{Box and whisker plot describing a relationship between a given response and its corresponding score. We check if the response lengths correlate with its scores.}
    \label{figure:length-bias-fc-unseen-combined}
\end{figure*}

\begin{figure*}[t!]
\centering
    \includegraphics[width=0.8\linewidth]{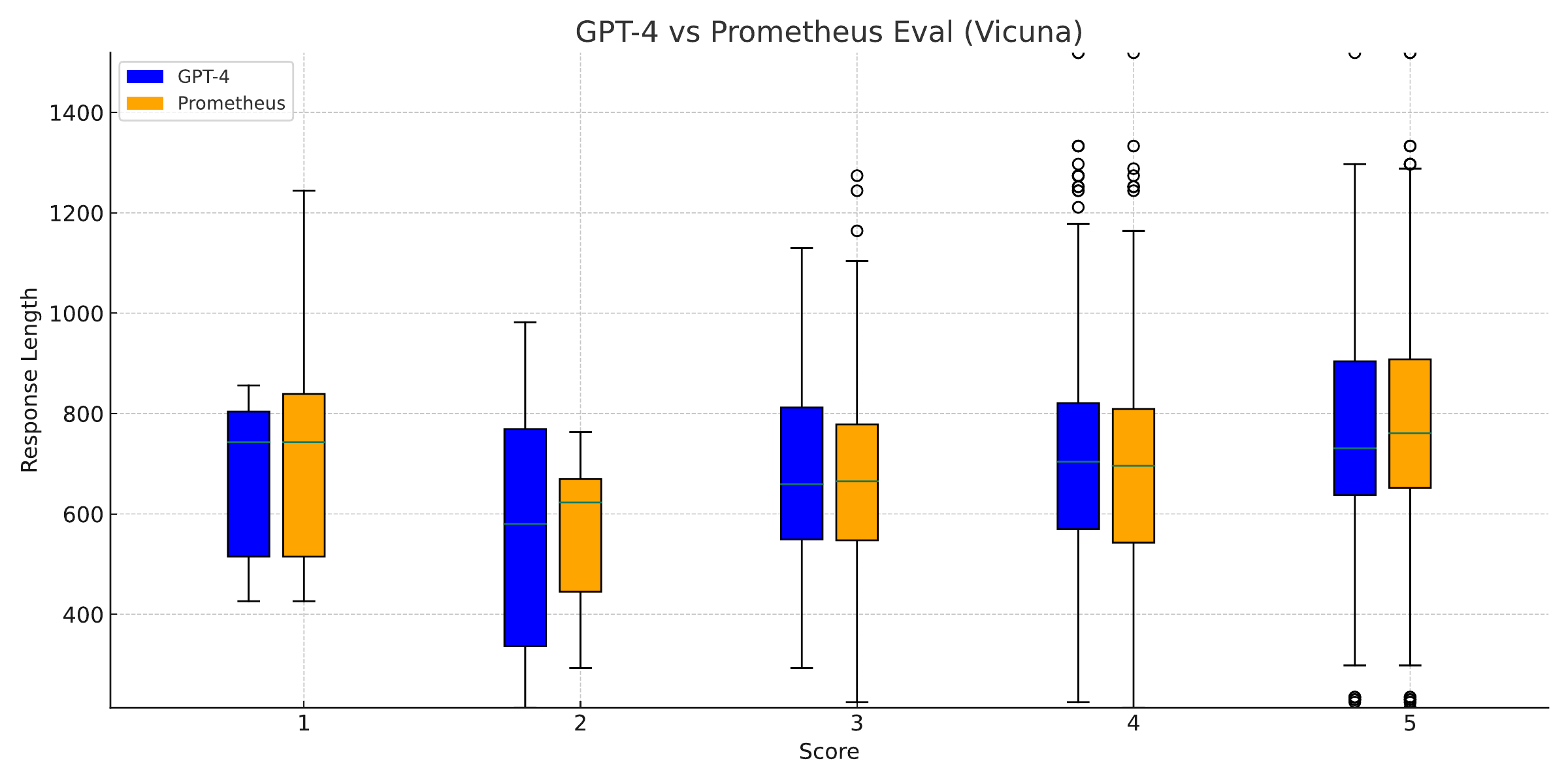}
    \caption{Box and whisker plot describing a relationship between a given response and its corresponding score. We check if the response lengths correlate with its scores.}
    \label{figure:length-bias-vicuna-combined}
\end{figure*}

One of the limitations of employing an LLM as an evaluator LM is that it could be vulnerable to various biases. In this work, we train/test on an Absolute Scoring evaluation setting, hence there exists no position bias. Yet, it is crucial to analyze whether \textsc{Prometheus} showed any bias towards favoring longer responses. Hence, we conduct a comprehensive analysis in this section.

As shown in Figure~\ref{figure:length-bias-fc-seen-combined}, Figure~\ref{figure:length-bias-fc-unseen-combined}, and Figure~\ref{figure:length-bias-vicuna-combined}, both GPT-4 and \textsc{Prometheus} and GPT-4 shows a similar trend of not favoring longer responses (i.e., similar length distribution among different scores). However, as mentioned in \citet{zheng2023judging}, LLM evaluators might favor more verbose responses, yet the responses from our test instances (\textsc{Feedback Bench} and Vicuna Bench) did not include any adversarial examples to test this phenomenon. More extensive research on whether the length bias is also transferred to fine-tuned evaluator LMs should be explored in future work.

\section{Prompt for \textsc{Feedback Collection} Creation}\label{appendix:prompt_augmentation}

In this section, we provide the extensive list of prompts used to create the \textsc{Feedback Collection}.

Note that in the prompt of generating a response and a feedback, we use the sentence length of the reference answer and append it to ``\{SENT NUM\}'' within the prompt. This was crucial to make the length even across different scores as shown in Figure~\ref{figure:length_bias}. Also, note that for the 1K score rubrics, 20K instructions \& reference answers, and 100K responses \& feedback within the \textsc{Feedback Collection}, each prompt was sequentially used. In early experiments, we found that generating every component all at once leads to very poor generation quality (i.e., similar responses \& feedback across different score ranges). Yet, we found that grouping (1) the instruction and reference answer generation and (2) the response and feedback generation had a positive synergy, leading to better generation quality and less amount of cost. Also, to the best of our knowledge, we are first to explore acquiring negative and neutral responses (Score 1 $\sim$ 4 responses) through GPT-4 augmentation. We hope future work could also explore applying this strategy to different use cases.

\begin{mybox}{Prompt for Brainstorming New Score Rubrics}
We are brainstorming criteria with which to grade a language model on its responses in diverse situations.\\
A `criteria` is some useful, real-world objective, and associated rubric for scores 1-5, that tests a capability.\\
\\
Here you will see 4 examples of `criteria`, and their scoring rubrics, formatted as JSON.\\
Criteria 1:\\
\{JSON LIST 1\}\\
\\
Criteria 2:\\
\{JSON LIST 2\}\\
\\
Criteria 3:\\
\{JSON LIST 3\}\\
\\
Criteria 4:\\
\{JSON LIST 4\}\\
\\
Please brainstorm a new criteria and scoring rubrics.\\
Be creative and create new but useful criteria that people in different settings or industries might find practical.\\
Please format the output as same as the above examples with no extra or surrounding text. Write [END] after you are done.\\
\\
New Criteria:
\end{mybox}

\begin{mybox}{Prompt for Paraphrasing as a New Score Rubric}
Please paraphrase the sentences inside the dictionary below.\\
Each paraphrase should not change the meaning or substance of the original sentence, be naturally written, but sufficiently diverse from one another.\\
Diversity can come from differences in diction, phrasing, sentence structure, formality, detail, and/or other stylistic changes.\\
\\
The dictionary:\\
\{CRITERIA\}\\
\\
Respond with only dictionary (same format as the given dictionary) with no extra or surrounding text.\\
Write [END] after you are done.\\
\\
Dictionary with Paraphrased Sentences:
\end{mybox}

\begin{mybox}{Prompt for Generating an Instruction and Reference Answer}
Your job is to generate a new novel problem and a response that is related to the given score rubric.\\
\\
The score rubric:\\
\{CRITERIA\}\\
\\
* Problem\\
- The problem should inherently be related to the score criteria and score rubric given above. Specifically, the score criteria should be the core attributes required to solve the problem.\\
- The problem itself should not be too generic or easy to solve.\\
- If the score rubric is related to logical abilities, generate problems that require math or coding abilities.\\
- Try to make the person who might solve the problem not notice the existence of the score rubric by not explicitly mentioning it, and also provide additional inputs and options if needed.\\
- Assume a situation where a user is interacting with an AI model. The user would try to ask in a first-person point of view, but not using terms like "I", "A User" or "You" in the first sentence.\\
- Do not give a role to the AI, assume that the user is asking a question from his point of view.\\
- Do not include any phrase related to AI model in the problem.\\
\\
* Response\\
- The response should be a response that would get a score of 5 from the score rubric.\\
- The response should be as detailed as possible unless the score rubric is related to conciseness or brevity. It should consist of multiple paragraphs, a list of items, or a step-by-step reasoning process.\\
- The response should look like how a well-prompted GPT-4 would normally answer your problem.\\
\\
* Format\\
- DO NOT WRITE ANY GREETING MESSAGES, just write the problem and response only.\\
- In front of the problem, append the phrase "Problem:" and in front of the response, append the phrase "Response:".\\
- Write in the order of "Problem" - "Response", where the two items are separated by the phrase "[NEXT]".\\
- Write [END] after you are done.\\
\\
Data Generation:\\
\end{mybox}

\begin{mybox}{Prompt for Generating Responses and Feedback}
Your job is to generate a response that would get a score of \{SCORE\} and corresponding feedback based on the given score rubric. For reference, a reference response that would get a score of 5 is also given.\\
\\
Instruction:\\
\{INSTRUCTION\}\\
\\
The score rubric:\\
\{CRITERIA\}\\
\\
Reference response (Score 5):\\
\{REFERENCE\}\\
\\
* Response\\
- The quality of the score \{SCORE\} response should be determined based on the score rubric, not by its length.\\
- The score \{SCORE\} response should have the same length as the reference response, composed of \{SENT NUM\} sentences.\\
- Do not explicitly state the keywords of the score rubric inside the response.\\
\\
* Feedback\\
- The score \{SCORE\} feedback should each be an explanation of why the response would get a score of \{SCORE\}. It should be written based on the generated response and score rubric.\\
- The score \{SCORE\} feedback shouldn't just copy and paste the score rubric, but it should also give very detailed feedback on the content of the corresponding response.\\
- The score \{SCORE\} feedback should include the phrase "So the overall score is \{SCORE\}" in the last sentence.\\
\\
* Format\\
- DO NOT WRITE ANY GREETING MESSAGES, just write the problem and response only.\\
- In front of the response, append the phrase "Response:" and in front of the feedback, append the phrase "Feedback:".\\
- Write in the order of "Response" - "Feedback", where the two items are separated by the phrase "[NEXT]".\\
- Write [END] after you are done.\\
\\
Data Generation:\\
\end{mybox}
\newpage

\section{Prompt used for \textsc{Prometheus}}\label{appendix:prompt_train_inference}

In this section, we provide the prompt used for training/inferencing \textsc{Prometheus}. Note that after applying the prompt template shown below, we also apply Llama-2's basic conversation prompt template in order to minimize the discrepancy between the training process of Llama-2 and training on the \textsc{Feedback Collection}.

\begin{mybox}{Prompt for Prometheus}
\#\#\#Task Description:\\
An instruction (might include an Input inside it), a response to evaluate, a reference answer that gets a score of 5, and a score rubric representing an evaluation criterion is given.\\
1. Write a detailed feedback that assesses the quality of the response strictly based on the given score rubric, not evaluating in general.\\
2. After writing a feedback, write a score that is an integer between 1 and 5. You should refer to the score rubric.\\
3. The output format should look as follows: \"Feedback: (write a feedback for criteria) [RESULT] (an integer number between 1 and 5)\"\\
4. Please do not generate any other opening, closing, and explanations.\\
\\
\#\#\#The instruction to evaluate:\\
\{instruction\}\\
\\
\#\#\#Response to evaluate:\\
\{response\}\\
\\
\#\#\#Reference Answer (Score 5):\\
\{reference answer\}\\
\\
\#\#\#Score Rubrics:\\
$[$\{ criteria description \}$]$\\
Score 1: \{score1 description\}\\
Score 2: \{score2 description\}\\
Score 3: \{score3 description\}\\
Score 4: \{score4 description\}\\
Score 5: \{score5 description\}\\
\\
\#\#\#Feedback: \\
\end{mybox}

\clearpage

\section{Qualitative Examples of Generated Feedback}\label{appendix:qualitative_example}

\begin{figure*}[t!]
\centering
    \includegraphics[width=1.0\linewidth]{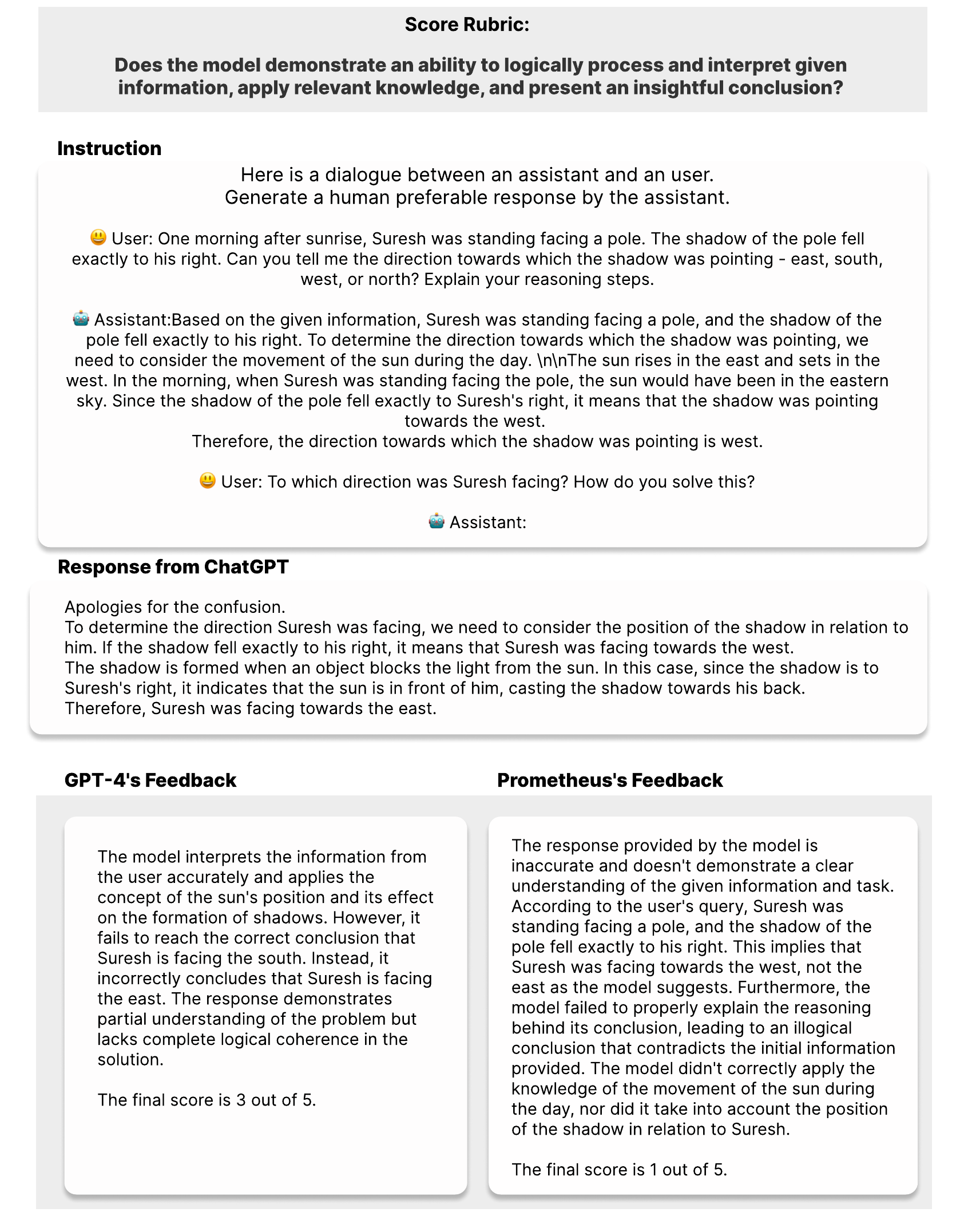}
    \caption{An example of comparing the feedback generated by \textsc{Prometheus} and GPT-4. GPT-4 was labeled to generate an abstract feedback.}
    \label{figure:too_abstract}
\end{figure*}

\begin{figure*}[t!]
\centering
    \includegraphics[width=1.0\linewidth]{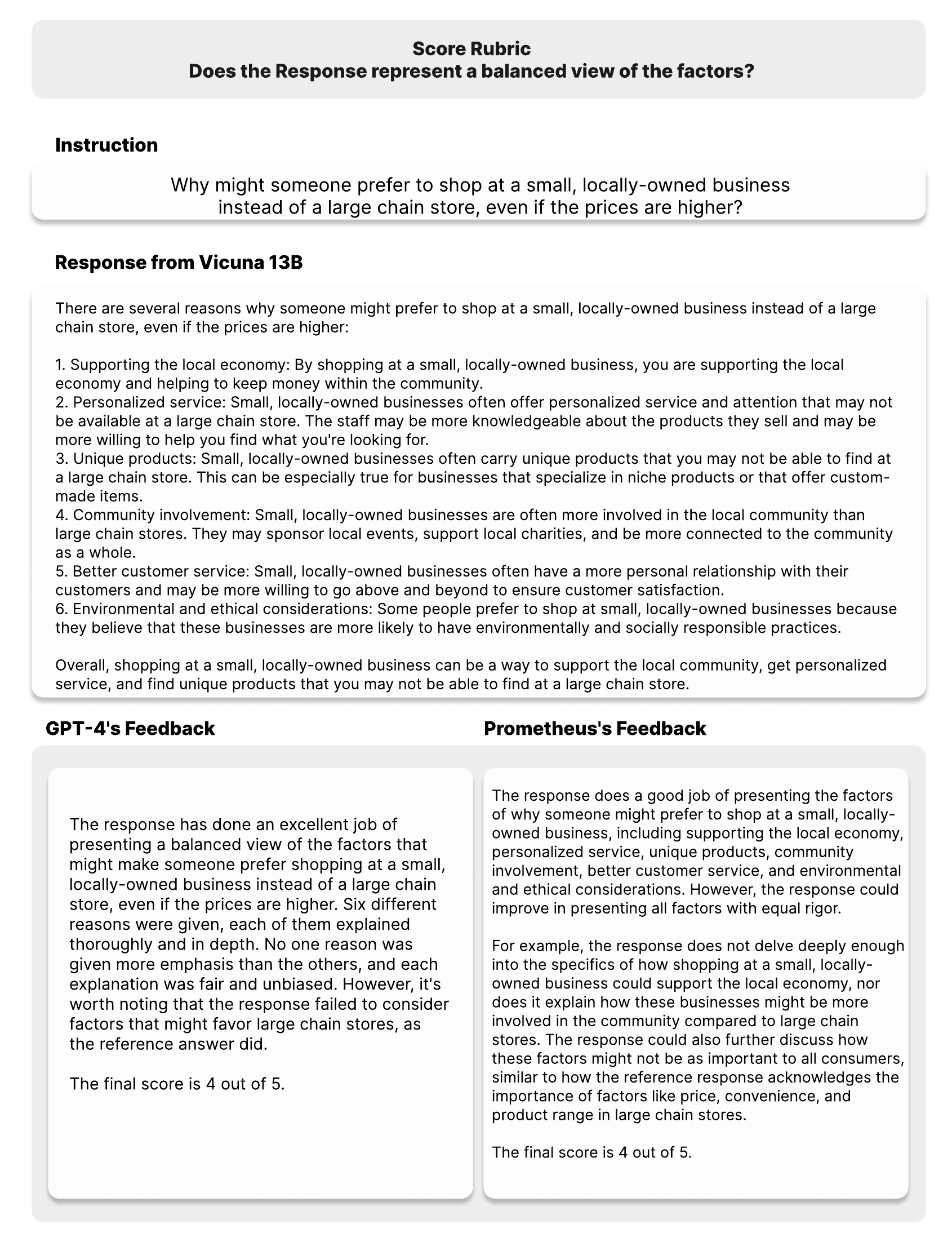}
    \caption{An example of comparing the feedback generated by \textsc{Prometheus} and GPT-4. \textsc{Prometheus} was labeled to generate an overly critical feedback.}
    \label{figure:over_critical}
\end{figure*}

\begin{figure*}[t!]
\centering
    \includegraphics[width=1.0\linewidth]{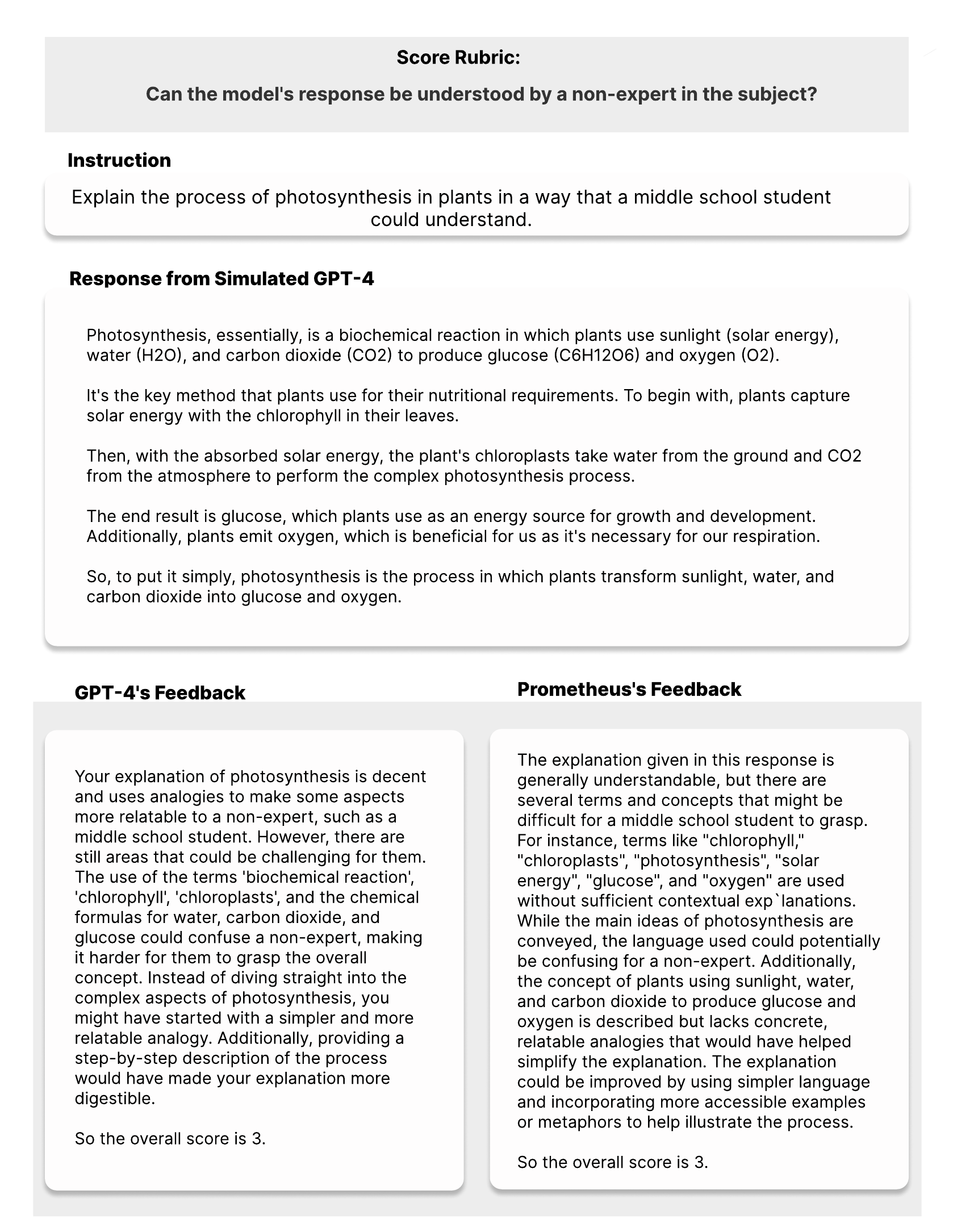}
    \caption{An example of comparing the feedback generated by \textsc{Prometheus} and GPT-4. Both evaluator LM generated a good feedback, and hence was labeled as a tie.}
    \label{figure:tie}
\end{figure*}

\begin{figure*}[t!]
\centering
    \includegraphics[width=1.0\linewidth]{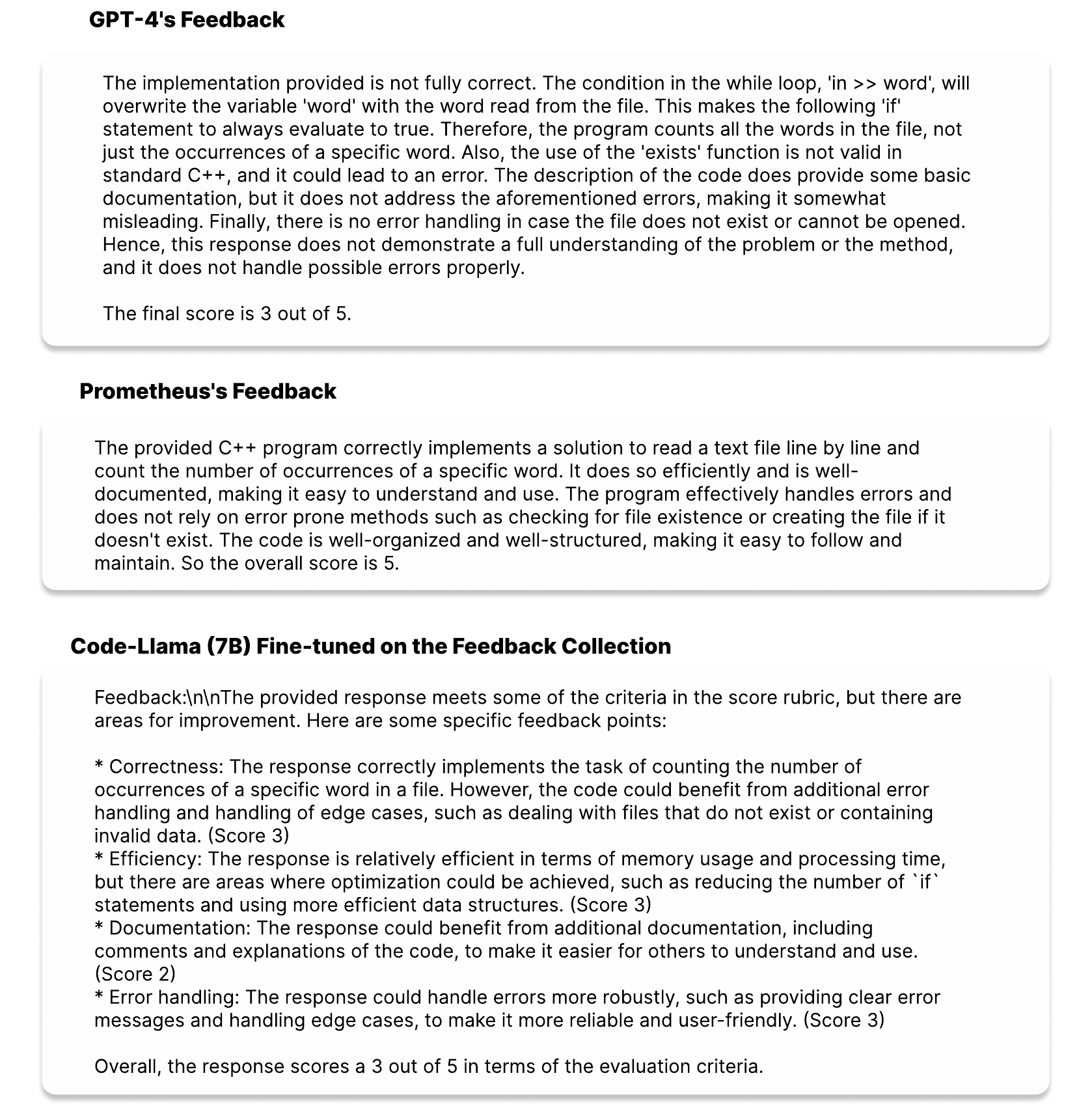}
    \caption{An example of comparing the feedback generated by \textsc{Prometheus}, GPT-4, and Code-Llama trained on the \textsc{Feedback Collection}. Compared to \textsc{Prometheus}, using a base model specialized on the code domain also helps to criticize and evaluate responses within the code domain.}
    \label{figure:code2}
\end{figure*}

Figure~\ref{figure:too_abstract}, Figure~\ref{figure:over_critical}, Figure~\ref{figure:tie}, Figure~\ref{figure:code2} shows a qualitative example of feedback generated by either GPT-4, \textsc{Prometheus} (13B), and Code-Llama trained on the \textsc{Feedback Collection}.

\section{Experimental Details for Human Evaluation}\label{appendix:human_eval_details_last}
\begin{figure*}[t!]
\centering
    \includegraphics[width=1.0\linewidth]{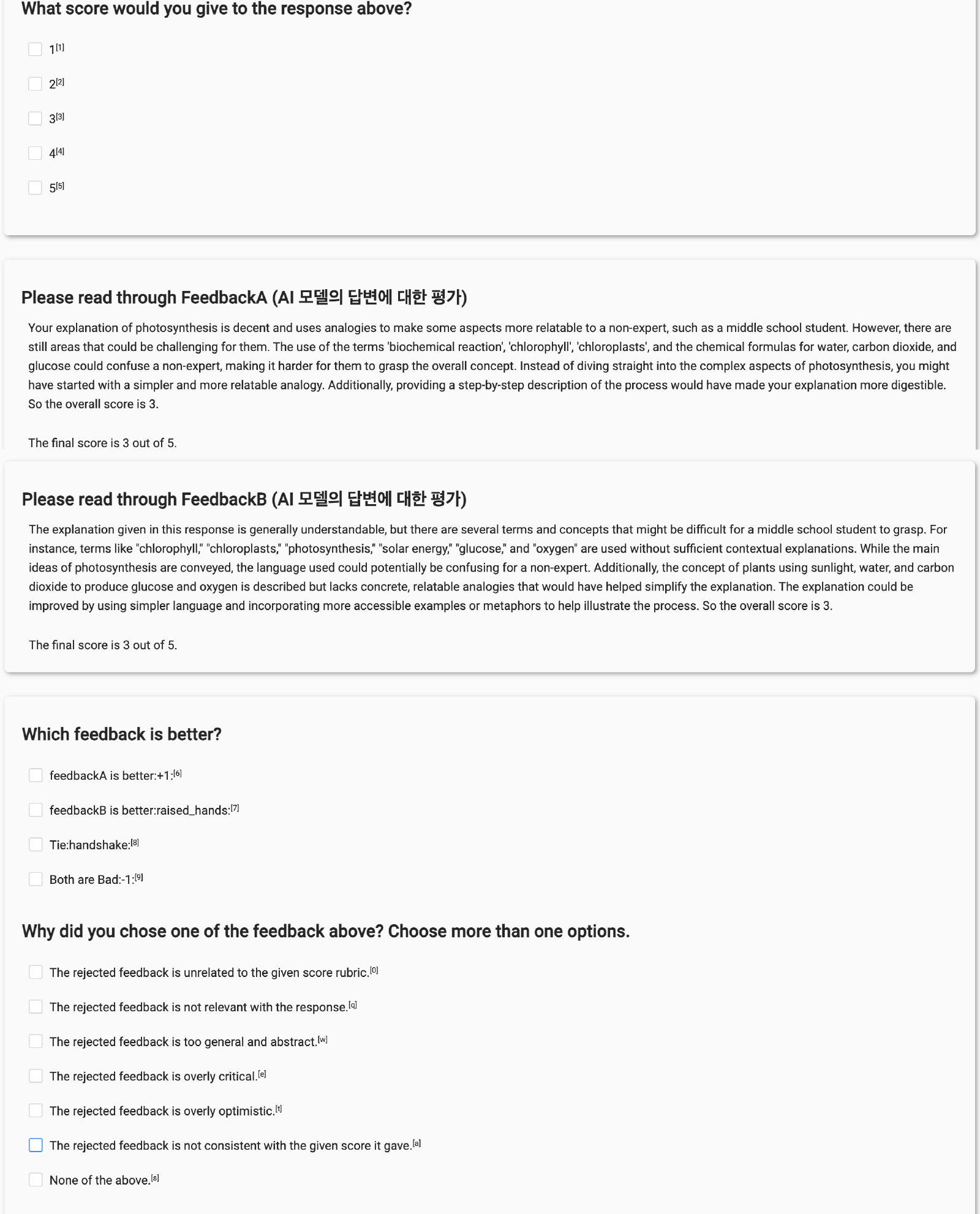}
    \caption{The annotation user interface for labeling the human scores, pairwise comparison of the two feedback and gathering labels of why one feedback was preferred over the other one.}
    \label{figure:annotation_ui2}
\end{figure*}

The user interface used for human evaluation is shown in Figure~\ref{figure:annotation_ui2}. In order to acquire a score decision for the response, a decision of which feedback is better, and an annotation of why they made a decision to choose one of the feedback, we constructed the user interface in sequential order. Each annotator was informed not to solve the questions below before solving the ones above them. The test instances were randomly selected among \textsc{Feedback Bench} (Customized Unseen Rubric set), Vicuna Bench, and MT Bench except for math and coding instances. The experiments were conducted on LabelStudio, hosted on Huggingface Spaces. We asked each annotator to spend at least 3 hours annotating 45 instances each and paid \$ 50. The 9 annotators were divided into 3 groups: (1) \textsc{Prometheus} \textit{vs} GPT-4, (2) \textsc{Prometheus} \textit{vs} GPT-3.5-Turbo, and (3) GPT-4 \textit{vs} GPT-3.5-Turbo. The order of the feedback was randomly decided in order to prevent order bias during annotation.

\clearpage

\begin{figure*}[t!]
\centering
    \includegraphics[width=1.0\linewidth]{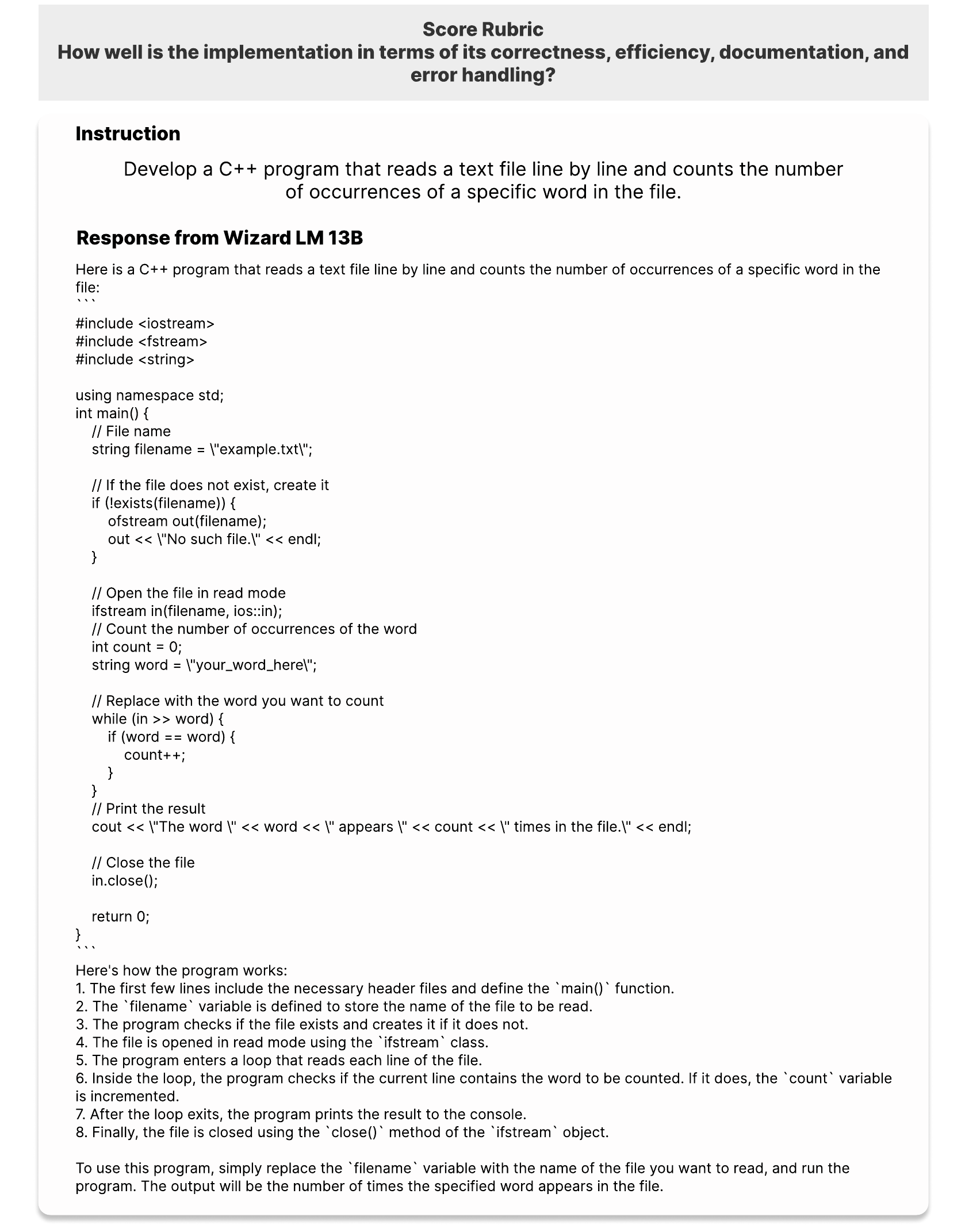}
    \label{figure:code1}
\end{figure*}

\begin{figure*}[t!]
\centering
    \includegraphics[width=1.0\linewidth]{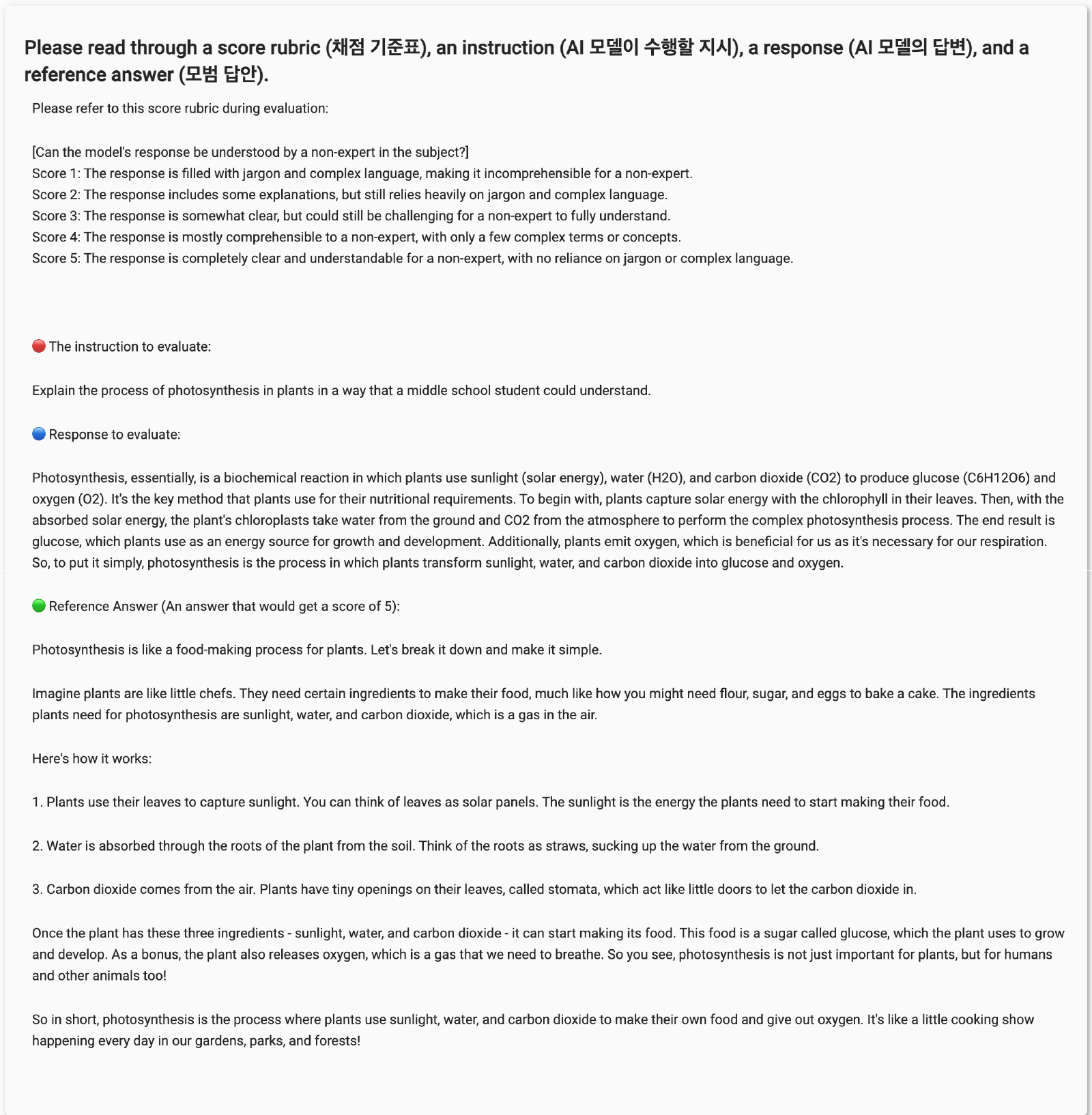}
    \label{figure:annotation_ui1}
\end{figure*}

\end{document}